\documentclass[11pt]{article}
\usepackage[utf8]{inputenc}
\usepackage[T1]{fontenc}
\usepackage{lmodern}
\usepackage{microtype}
\ifdefined\DeclareUnicodeCharacter
  \DeclareUnicodeCharacter{2014}{---}
  \DeclareUnicodeCharacter{2192}{\ensuremath{\to}}
\fi
\usepackage[margin=1in]{geometry}
\usepackage{booktabs}
\usepackage{graphicx}
\usepackage{float}
\usepackage{afterpage}
\usepackage[section]{placeins}

\usepackage{amsmath}
\usepackage{amssymb}
\usepackage{hyperref}
\usepackage{xcolor}
\usepackage{listings}
\definecolor{kwblue}{rgb}{0.10,0.20,0.65}
\definecolor{typepurple}{rgb}{0.50,0.15,0.55}
\definecolor{strred}{rgb}{0.62,0.12,0.12}
\definecolor{cmtgreen}{rgb}{0.20,0.45,0.20}
\lstdefinestyle{housespecyaml}{
  basicstyle=\ttfamily\footnotesize,
  backgroundcolor=\color{black!3},
  frame=single,
  framesep=5pt,
  rulecolor=\color{black!45},
  columns=fullflexible,
  breaklines=true,
  keepspaces=true,
  showstringspaces=false,
  alsoletter=_,
  commentstyle=\color{cmtgreen}\itshape,
  stringstyle=\color{strred},
  morecomment=[l]{\#},
  morestring=[b]",
  morestring=[b]',
  keywordstyle=\color{kwblue}\bfseries,
  keywords={World,Agent,Objective,Dial,Tuner,Judge,Simulation,def,return,elif,if,and,lambda},
  keywordstyle=[2]\color{typepurple}\bfseries,
  keywords=[2]{compile,sweep,search,refine,run,choose,score_trajectory,transition},
}
\lstdefinestyle{housespecyamlbare}{style=housespecyaml, frame=none,
  backgroundcolor=\color{white}, xleftmargin=0pt, framexleftmargin=0pt,
  aboveskip=1pt, belowskip=0pt}
\usepackage[most]{tcolorbox}
\tcbuselibrary{listings,skins}
\usepackage{tikz}
\usetikzlibrary{positioning,shapes.geometric,arrows.meta,calc,fit,backgrounds,shadows}
\definecolor{cardindigo}{HTML}{4F46E5}
\definecolor{slate}{HTML}{475569}
\definecolor{amber}{HTML}{D97706}
\definecolor{cardblue}{HTML}{1D4ED8}
\definecolor{cardpurple}{HTML}{6D28D9}
\definecolor{cardteal}{HTML}{0F766E}
\tcbset{
  cardcol/.store in=\cardcol, cardcol=cardblue,
  policycard/.style={
    enhanced, breakable=false, arc=3pt, boxrule=0.7pt, top=3pt, bottom=3pt,
    left=7pt, right=7pt, boxsep=1pt, coltitle=white,
    colframe=\cardcol!60!black, colback=white, colbacktitle=\cardcol,
    fonttitle=\bfseries\small, drop fuzzy shadow=black!25,
  },
}
\usepackage{url}
\usepackage{multirow}
\usepackage{array}
\usepackage{longtable}
\usepackage{ragged2e}
\newcolumntype{L}[1]{>{\RaggedRight\arraybackslash}p{#1}}
\usepackage[htt]{hyphenat}
\hypersetup{colorlinks=true, linkcolor=blue, citecolor=blue, urlcolor=blue}

\usepackage{caption}
\DeclareCaptionLabelFormat{boldunder}{\underline{\textbf{#1~#2}}}
\let\oldincludegraphics\includegraphics
\renewcommand{\includegraphics}[2][]{%
  \setlength{\fboxsep}{0pt}\setlength{\fboxrule}{0.5pt}%
  \fbox{\oldincludegraphics[#1]{#2}}}

\IfFileExists{numbers.tex}{
\newcommand{\CodeFirstDivergence}{21}
\newcommand{\LLMFirstDivergence}{2.3}
\newcommand{\LLMFinalLOne}{10.3}
\newcommand{\RolloutSteps}{20}

\newcommand{\CodeStepsPerSec}{14,997}
\newcommand{\LLMStepsPerSec}{0.32}

\newcommand{\CodeOODRate}{100\%}

\newcommand{\LLMProbeErrorRate}{60\%}

\newcommand{\JudgeSpearman}{0.75}

\newcommand{\TuningBudget}{200}
\newcommand{\NumTasks}{20}
\newcommand{\NumExperiments}{79}
\newcommand{\WTNumTrainWorlds}{60}
\newcommand{\WTNumTestWorlds}{20}
\newcommand{\WTNumCases}{800}

\newcommand{\WTSmallName}{0.5B}
\newcommand{\WTSmallBase}{37\%}
\newcommand{\WTSmallFT}{66\%}
\newcommand{\WTSmallGain}{29}
\newcommand{\WTLargeName}{32B}
\newcommand{\WTLargeBase}{78\%}
\newcommand{\WTLargeFT}{81\%}
\newcommand{\WTLargeGain}{3}
\newcommand{\WTOracle}{86\%}
\newcommand{\WTFloor}{34\%}
\newcommand{\WTMetaFewShot}{70\%}
\newcommand{\WTScratchFewShot}{68\%}
\newcommand{\WCBase}{48\%}
\newcommand{\WCMaxWorlds}{512}
\newcommand{\WCMaxAcc}{59\%}
\newcommand{\WCMaxGain}{10}
\newcommand{\WCOracle}{69\%}
\newcommand{\CodeSynBaseFive}{84\%}
\newcommand{\CodeSynFtFive}{95\%}
\newcommand{\CodeHeBaseOne}{78\%}
\newcommand{\CodeHeFtOne}{70\%}
\newcommand{\CodeHeBaseFive}{87\%}
\newcommand{\CodeHeFtFive}{91\%}
\newcommand{\CodeNumWorlds}{164}
\newcommand{\ArcNumTrainWorlds}{400}
\newcommand{\ArcNumEval}{100}
\newcommand{\ArcNumTTT}{40}
\newcommand{\ArcZeroShot}{2\%}
\newcommand{\ArcTTTLight}{8\%}
\newcommand{\ArcTTTHeavy}{10\%}
\newcommand{\ArcTTTCorrupt}{2\%}
\newcommand{\ArcTTTGain}{8}
\newcommand{\ArcLadderBase}{3\%}
\newcommand{\ArcLadderTop}{5\%}
\newcommand{\ArcLadderTopN}{200}
\newcommand{\ArcHeavyCI}{[2, 20]}
\newcommand{\CrossWorldNTrain}{128}
\newcommand{\CrossWorldNEval}{60}
\newcommand{\CrossWorldNSeeds}{3}
\newcommand{\CrossWorldBase}{0\%}
\newcommand{\CrossWorldMean}{40\%}
\newcommand{\CrossWorldCI}{[35, 45]}
\newcommand{\CrossWorldCorrupt}{6\%}
\newcommand{\CrossWorldCorruptCI}{[4, 10]}
\newcommand{\CrossWorldGap}{34}
\newcommand{\CrossWorldGapCI}{[29, 39]}
\newcommand{\CrossWorldSeeds}{42\%, 43\%, 36\%}
\newcommand{\CrossWorldLadderLo}{29\%}
\newcommand{\CrossWorldLadderHi}{49\%}
\newcommand{\CrossWorldLadderLoN}{16}
\newcommand{\ArcSeedsN}{2}
\newcommand{\ArcHeavySeedMean}{11\%}
\newcommand{\ArcHeavySeedList}{10\%, 12\%}
\newcommand{\ArcCorruptSeedMean}{4\%}

\newcommand{\TrajNSeeds}{3}
\newcommand{\TrajTrainTraj}{70\%}
\newcommand{\TrajTrainAnswer}{58\%}
\newcommand{\TrajTrainGap}{12}
\newcommand{\TrajLongTraj}{68\%}
\newcommand{\TrajLongAnswer}{75\%}
\newcommand{\LfNumWorlds}{60}
\newcommand{\LfZero}{35\%}
\newcommand{\LfLight}{66\%}
\newcommand{\LfHeavy}{69\%}
\newcommand{\LfCorrupt}{3\%}
\newcommand{\LfHeavyCI}{[59, 77]}
\newcommand{\ClrsNumWorlds}{29}
\newcommand{\ClrsZero}{9\%}
\newcommand{\ClrsLight}{13\%}
\newcommand{\ClrsHeavy}{17\%}
\newcommand{\ClrsCorrupt}{4\%}
\newcommand{\BongNumWorlds}{221}
\newcommand{\BongBase}{46\%}
\newcommand{\BongHeavy}{48\%}
\newcommand{\BongCorrupt}{52\%}
\newcommand{\ComboNWorlds}{79}

\newcommand{\LawNWorlds}{333}
\newcommand{\LawNDomains}{5}
\newcommand{\LawLodoRtwo}{0.51}
\newcommand{\LawWithinArc}{0.98}
\newcommand{\LawWithinListfn}{0.90}
\newcommand{\LawWithinClrs}{0.73}
\newcommand{\LawWithinBongard}{0.77}
\newcommand{\LawWithinTabular}{0.63}
\newcommand{\LawTabOOS}{0.60}
\newcommand{\LawFreeProxyRtwo}{-0.38}
\newcommand{\SymScaleLo}{0.04}
\newcommand{\SymScaleHi}{0.16}
\newcommand{\SymCorrupt}{0.00}
\newcommand{\FrameNWorlds}{20}
\newcommand{\FrameNLangs}{5}
\newcommand{\FrameLangs}{Python, C++, Java, JS, Go}
\newcommand{\FrameTttSteps}{60}
\newcommand{\FrameQwenZero}{28\%}
\newcommand{\FrameQwenTtt}{30\%}

\newcommand{\FrameLlamaZero}{29\%}
\newcommand{\FrameLlamaTtt}{27\%}
\newcommand{\FrameLlamaCorrupt}{24\%}

\newcommand{\HybridNTrain}{80}
\newcommand{\HybridNEval}{40}
\newcommand{\HybridRetry}{5}
\newcommand{\HybridQwenPairs}{76}
\newcommand{\HybridQwenBase}{80\%}
\newcommand{\HybridQwenAmort}{85\%}
\newcommand{\HybridQwenHybrid}{90\%}
\newcommand{\HybridQwenAmortDelta}{+5}
\newcommand{\HybridQwenHybridDelta}{+10}
\newcommand{\HybridQwenCalls}{1.45}
\newcommand{\HybridLlamaPairs}{66}
\newcommand{\HybridLlamaBase}{57\%}
\newcommand{\HybridLlamaAmort}{50\%}
\newcommand{\HybridLlamaHybrid}{65\%}
\newcommand{\HybridLlamaAmortDelta}{-7}
\newcommand{\HybridLlamaHybridDelta}{+8}
\newcommand{\HybridLlamaCalls}{2.65}
\newcommand{\WCListfn}{3}
\newcommand{\WCArc}{9}
\newcommand{\WCClrs}{15}
\newcommand{\CartOpenWorldSolve}{100\%}

\newcommand{\ProtoNumWorlds}{100}

\newcommand{\MGOpenWorldPlan}{11}
\newcommand{\MGDreamerFirstSolve}{9{,}640}

\newcommand{\NumModels}{9}
\newcommand{\ModelList}{\texttt{gemma2:9b}, \texttt{gpt-oss:20b}, \texttt{llama3.1:8b}, \texttt{qwen2.5:1.5b}, \texttt{qwen2.5:3b}, \texttt{qwen2.5:7b}, \texttt{qwen2.5:14b}, \texttt{qwen2.5-coder:7b}, \texttt{qwen3-coder:30b}}
\newcommand{\MultiCodeExact}{24/24}
\newcommand{\MultiWorldRollouts}{24}
\newcommand{\MultiCodeCI}{[0.86, 1.00]}
\newcommand{\MultiLLMExact}{0/24}
\newcommand{\MultiLLMCI}{[0.00, 0.14]}
\newcommand{\MLPInDistTenK}{50\%}
\newcommand{\MLPOODTenK}{0\%}
\newcommand{\KNNInDistTenK}{30\%}
\newcommand{\KNNRolloutsTenK}{8/8}

\newcommand{\OrderConsistency}{28\%}

\newcommand{\RubricParaRho}{0.58}

\newcommand{\PooledMcNemarP}{0.078}

\newcommand{\CodeHundredX}{100\%}
\newcommand{\DeltaMLPInDist}{50\%}
\newcommand{\DeltaMLPHundredX}{0\%}

\newcommand{\CplxFour}{1.00}
\newcommand{\CplxEight}{0.53}
\newcommand{\CplxTwelve}{0.25}
\newcommand{\CplxSixteen}{0.15}
\newcommand{\StochAccept}{67\%}
\newcommand{\StochArrivalErr}{1.2}
\newcommand{\StochDetAcc}{100\%}
\newcommand{\StochOracleExact}{100\%}
\newcommand{\PlanCode}{7.5}
\newcommand{\PlanOracle}{7.5}
\newcommand{\PlanLLM}{0.0}
\newcommand{\PlanReactive}{5.0}
\newcommand{\PlanRandom}{3.8}
\newcommand{\PlanLLMSeconds}{86}
\newcommand{\PlanCodeSeconds}{0.03}
\newcommand{\SelfCheckPrograms}{30}
\newcommand{\SelfCheckPairs}{900}
\newcommand{\SelfCheckPrecision}{100\%}
\newcommand{\SelfCheckRecall}{100\%}
\newcommand{\SelfCheckSpearman}{1.00}

\newcommand{\SweAtomicN}{6}
\newcommand{\SweStagedN}{15}
\newcommand{\SweBudget}{4}

\newcommand{\CompRules}{16}
\newcommand{\CompChildRules}{4}
\newcommand{\NestSteps}{20}

\newcommand{\IndRulesIn}{1.00}

\newcommand{\IndTracesIn}{0.43}
\newcommand{\IndTracesOod}{0.43}

\newcommand{\IndMlpOod}{0.00}
\newcommand{\IndKnnOod}{0.00}
\newcommand{\ScaleQwenSmall}{0.43}
\newcommand{\ScaleQwenCoder}{0.33}
\newcommand{\ScaleGptoss}{0.43}
\newcommand{\ScaleNumModels}{3}

\newcommand{\ActiveCands}{108}
\newcommand{\ActiveRules}{5}

\newcommand{\ActiveMeanSteps}{11.2}
\newcommand{\ActivePassiveSteps}{14.7}
\newcommand{\ActiveClairSteps}{11.6}
\newcommand{\ActivePassiveUnresolved}{2}

\newcommand{\BakeMethods}{6}
\newcommand{\BakeDomains}{2}
\newcommand{\BakeCodeReturn}{7.5}
\newcommand{\BakeLearnedOodMax}{0.00}
\newcommand{\BakeCodeSpeed}{2{,}250{,}984}

\newcommand{\AblVerified}{35\%}

\newcommand{\AblNoisyFull}{19\%}

\newcommand{\AblFloorHard}{23\%}
\newcommand{\AblWorlds}{256}
\newcommand{\AblSeeds}{3}

}{}

\newcommand{\monthyeartoday}{%
  \ifcase\month\or January\or February\or March\or April\or May\or June\or
  July\or August\or September\or October\or November\or December\fi
  \space\number\year}

\newcommand{\openworld}{\textsc{OpenWorld}}

\begin{document}

\thispagestyle{empty}
\noindent
\begin{minipage}[c]{0.5\textwidth}
  \raggedright\oldincludegraphics[height=0.8cm]{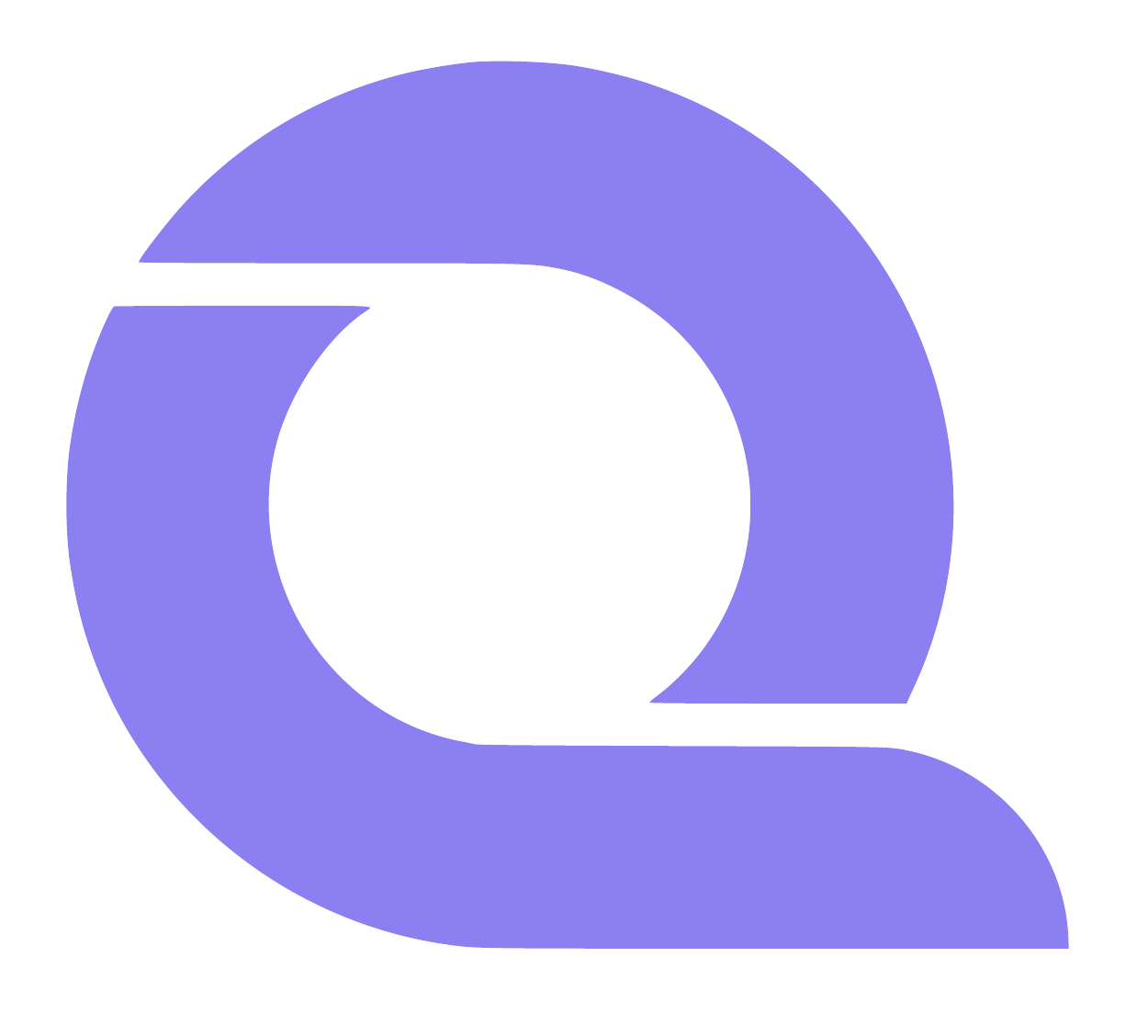}
\end{minipage}\hfill
\begin{minipage}[c]{0.5\textwidth}
  \raggedleft\itshape \monthyeartoday
\end{minipage}

\vspace{0.7em}
\noindent\oldincludegraphics[width=\textwidth]{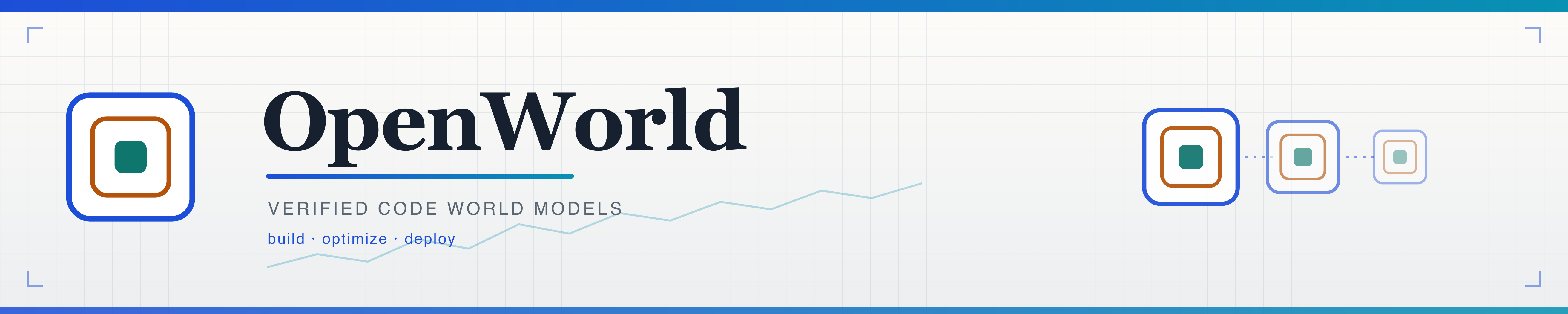}

\vspace{0.55em}
\begin{center}
  {\LARGE\bfseries World-Time Compute with\\[2pt]
   Verified Code World Models\par}

  \vspace{0.5em}
  {\color{cardblue}\rule{0.3\textwidth}{1.2pt}}

  \vspace{0.55em}
  {\normalsize
   \mbox{James Schwoebel\textsuperscript{1,\,*}}~\textcolor{cardblue!80}{\textbullet}
   \mbox{Ingrida Semenec, PhD\textsuperscript{1}}~\textcolor{cardblue!80}{\textbullet}
   \mbox{Jenia Rousseva, PhD\textsuperscript{1}}~\textcolor{cardblue!80}{\textbullet}
   \mbox{Marcos Ortiz, PhD\textsuperscript{1}}~\textcolor{cardblue!80}{\textbullet}
   \mbox{Collin Overbay\textsuperscript{1}}~\textcolor{cardblue!80}{\textbullet}
   \mbox{Christopher Klaus, PhD\textsuperscript{7}}~\textcolor{cardblue!80}{\textbullet}
   \mbox{Anderson Edmond\textsuperscript{8}}~\textcolor{cardblue!80}{\textbullet}
   \mbox{Manish Bhatt\textsuperscript{4,\,6}}~\textcolor{cardblue!80}{\textbullet}
   \mbox{Rome Thorstenson\textsuperscript{3}}~\textcolor{cardblue!80}{\textbullet}
   \mbox{Jessica Tsai, MD, PhD\textsuperscript{1}}~\textcolor{cardblue!80}{\textbullet}
   \mbox{Martin G. Frasch, MD, PhD\textsuperscript{2,\,5}}\par}

  \vspace{0.5em}
  {\footnotesize\color{black!70}
   \textsuperscript{1}\,Quome, Inc.\quad
   \textsuperscript{2}\,University of Washington\quad
   \textsuperscript{3}\,Rafter Labs, Inc.\quad
   \textsuperscript{4}\,OWASP\quad
   \textsuperscript{5}\,Health Stream Analytics, LLC\quad
   \textsuperscript{6}\,IEEE Computer Society\quad \textsuperscript{7}\,Fusen World LLC\quad \textsuperscript{8}\,Independent\quad\textsuperscript{*}\,Corresponding author\par}
\end{center}

\vspace{0.15em}
\begin{tcolorbox}[breakable, enhanced, colback=white, colframe=black!12,
  arc=3pt, boxrule=0.5pt, left=18pt, right=16pt, top=5pt, bottom=5pt,
  borderline west={3pt}{0pt}{cardblue}, drop fuzzy shadow=black!18]
\setlength{\parskip}{0pt}
{\sffamily\bfseries\color{cardblue} ABSTRACT}\par
\smallskip
\fontsize{9.0}{10.7}\selectfont
A large language model learns to generalize across a domain only after it sees many real,
labeled examples from that domain, and in most domains those examples are scarce. We study a
cheap way to make them. When a domain's dynamics can be written as code, one template turns
into many \emph{world models} (simulators whose dynamics are executable,
verifiable programs over symbolic state), and each one is an endless source of \emph{exactly}%
-labeled trajectories. We fine-tune an LLM on trajectories run through many such world
models of a domain (\textbf{world-time compute}, the training-time cousin of
test-time compute), and this raises how well it generalizes to \emph{held-out} worlds it never trained on (shown on
synthesized world families). The gains are largest where the model is weakest
($+\WTSmallGain$ points at \WTSmallName; the largest model's lift is within noise, which fits
task saturation). You can trust the labels because the worlds are \emph{verified code}, not
learned approximations: dynamics that are synthesized and then checked stay exact over
\RolloutSteps-step rollouts and answer $10\times$ out-of-distribution probes exactly
(\CodeOODRate), while per-step LLM prediction and a trained MLP pile up error and
collapse. Unlike domain randomization (which varies one simulator's parameters), we author
and verify each world on its own; a corrupted-label control shows that label \emph{exactness}, not
task variety, drives the held-out gains.

\smallskip\noindent
On real, human-authored benchmarks (ARC-AGI grids, List Functions, CLRS
algorithms) the same lever works as \emph{per-world} test-time training: it grows with compute
and collapses under label corruption. On one coherent real family (List Functions) the harder
\emph{cross-world} form works too: a single adapter trained on \CrossWorldNTrain{} \emph{disjoint}
worlds generalizes to held-out worlds it never saw, reaching \CrossWorldMean{} against \CrossWorldCorrupt{}
for a corrupted-label control (a \mbox{+\CrossWorldGap}-point exactness-gated lift, CI
\CrossWorldGapCI). The gain follows a \emph{descriptive} regularity---a
curve that saturates, not a law: it is largest for \emph{few-step} reasoning and small, weak
models, and it fades for long reasoning chains, for tasks read off rich perception, and as a
task saturates; transfer across tasks is weak when worlds share no skill.
The worlds are authored and served by \textbf{\openworld}, a zero-dependency framework that makes
it cheap to assemble the \emph{many} worlds a domain needs (the framework is a companion paper).
\textbf{Scope: the symbolic-state regime}: where you can write down the dynamics, verified code makes a
\emph{correct} world model a zero-training prototype; pixel-native domains stay the territory of
learned models. All code, recipes, and this manuscript regenerate from one public repository.

\smallskip
\noindent\textbf{Keywords:}~%
world models; code world models; neuro-symbolic AI; program synthesis;
value alignment; agents-as-a-judge; local LLMs.

\smallskip
\noindent\textcolor{blue}{\textbf{Code \& data:}}~\href{https://github.com/quome-cloud/openworld}{\texttt{github.com/quome-cloud/openworld}}\quad\textcolor{blue}{\textbf{Correspondence:}}~\href{mailto:jim@quome.com}{jim@quome.com}
\end{tcolorbox}
\clearpage

\section{Introduction}
\begin{figure}[t]
\centering
\includegraphics[width=0.97\linewidth]{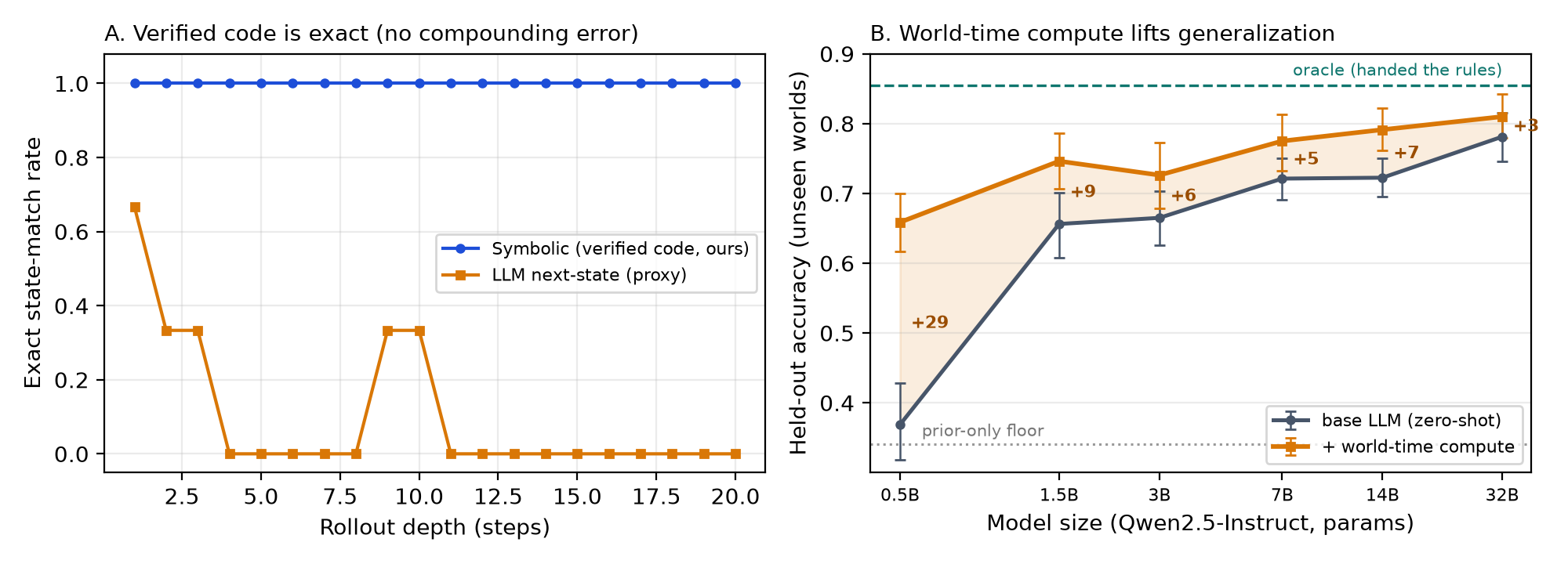}
\caption{\emph{A:} The task is to predict a world's next symbolic state from the current
state and an action (here the sprint backlog/debt/bug world, wired to a
ground-truth oracle). Dynamics synthesized and verified once as
code (blue) match the oracle exactly at every depth of a \RolloutSteps-step
rollout, while per-step LLM next-state prediction (orange)---a structural
stand-in for learned neural dynamics---first goes wrong at step
\LLMFirstDivergence{} on average and never finishes an exact rollout. (Panel A is the
per-depth exact-match rate over $n{=}3$ rollouts, so the orange trace is noisy and
non-monotonic; the solid claim is the contrast with code's exact-at-every-depth line, not
the shape of the orange.) Programs encode rules, not statistics, so the compounding-error failure of learned
world models is ruled out by construction, at zero training cost. \emph{B
(payoff):} what that exactness buys---world-time compute. Fine-tuning
an LLM on trajectories run through \WTNumTrainWorlds{} verified
train world models raises held-out diagnostic accuracy on
\WTNumTestWorlds{} unseen worlds toward the rules-given oracle, with the
largest gain for the smallest models ($+\WTSmallGain$ points at \WTSmallName,
shrinking to $+\WTLargeGain$ at \WTLargeName)---a generalization lever that is
strongest where capability is scarcest (bars: 95\% bootstrap CIs over held-out
specialties; dotted line: the prior-only floor; E74, \S\ref{sec:world-time}).}
\label{fig:hero}
\end{figure}

\textbf{Manufacturing experience.} A language model learns to generalize across a domain only after
it sees many real, labeled examples of it---and in most domains those examples are scarce and
costly. We study a cheap way to make them. When a domain's dynamics can be written as
code, a single template turns into many \emph{world models}---simulators whose
dynamics are executable, verifiable programs over symbolic state---and each one
generates exactly-labeled trajectories without end. Fine-tuning a model on trajectories
run through many such worlds of a domain---\emph{world-time compute}
(\S\ref{sec:world-time}), the training-time analogue of test-time compute---raises how well it
generalizes to held-out worlds it never trained on. Two things have to hold.
First, the worlds must not lie: a world teaches
the right rules only if its dynamics are correct, so we build them as verified code
rather than learned approximations. A model writes the transition as a program, and we accept it only after it
parses with the required signature, runs in a sandbox without breaking the world's declared
invariants, and (optionally) survives a second model's critique against the rules. The accepted
program stays exact over long rollouts with no compounding error
(Figure~\ref{fig:hero}A). Second, world-time compute needs many worlds, and that is
practical only if each one is cheap to build. The tool that makes it cheap is \openworld{}, a
zero-dependency framework that turns authoring a verified world model
(perceive\,$\to$\,world\,$\to$\,emit\,$\to$\,act, served and shareable) into a minutes-scale,
push-button task (the companion framework paper~\cite{openworld_framework}). No general tool
has made authoring the many verified world models this needs cheap; \openworld{} is that
tool, and we use it here to manufacture training experience.

A world model is an agent's internal simulator: given a state and an action, it
predicts the next state (and/or observation), which lets the agent plan ``in
imagination'' instead of by trial in the real world~\cite{ha2018world}. The whole field shares that
\emph{genus}; where research programs split is \emph{how the
model represents state and how it is obtained}, and today the term names two largely
separate \emph{species}. The dominant, most visible one is \emph{perceptual and
generative}: a neural network learned from large amounts of data, whose state is a
continuous latent (or pixels, video, or a 3D scene) and whose output is a
\emph{generated observation}. It runs from the V--M--C triad of Ha and
Schmidhuber~\cite{ha2018world} through the RSSM-based Dreamer
family~\cite{dreamerv3}, value-equivalent planning in MuZero~\cite{muzero},
autoregressive token models such as IRIS~\cite{iris2023} and
$\Delta$-IRIS~\cite{deltairis2024}, diffusion dynamics in
DIAMOND~\cite{diamond2024}, and joint-embedding predictive video models that
forecast in representation space such as V-JEPA~\cite{vjepa2025}. Most
visibly in recent generative AI, the species now reaches interactive generative
environments and ``spatial intelligence''---internet-scale video and 3D world
generators such as Genie~\cite{genie2024}, video-generation-as-simulator
(Sora)~\cite{sora2024}, and Fei-Fei Li's World Labs~\cite{worldlabs2025}. Powerful as these are, they carry three built-in
costs: they are \emph{data-hungry} (millions of environment steps, or
internet-scale video), \emph{compounding} (every learned transition adds error
that builds up over a rollout), and \emph{opaque} (both dynamics and values
live in continuous weights you cannot inspect, edit, or audit).

\paragraph{What ``world model'' means in this paper.} The term has come to
suggest generative, pixel- and 3D-native systems, but we use a simpler, narrower sense---a
forward predictor of
the next \emph{state}, for planning---and we hold the state to a \emph{symbolic}
object: a typed record of named variables (counts, prices, positions, flags), not
pixels, latents, or 3D geometry. Our world models are therefore the second species,
\emph{Code World Models} (CWMs): the transition is an explicit, human-readable
\emph{program} over that symbolic state, synthesized by an LLM from a declared rule
set and verified before use, and a rollout returns the next symbolic state
exactly rather than a generated image, with zero training data. This
is the same genus as the perceptual models above (a simulator the agent rolls
forward) but a different species---and one we mean to be complementary.
OpenWorld aims at domains you can \emph{specify} symbolically (inventories,
markets, schedules, negotiations, physics with known equations); it is not a
pixel-native, perceptual, or open-world-3D model, and it does not generate images,
video, or scenes. Where the spatial-intelligence systems \emph{learn what cannot be
told}, OpenWorld \emph{writes down what can be told, and verifies it}.

\paragraph{World, action, agent, simulation.} The world model is only one of four separate
objects. A \emph{world} is the
environment-as-simulator: a symbolic state, a discrete set of \emph{actions}, and
the verified transition that maps a $(\textrm{state}, \textrm{action})$ pair to the
next state (plus optional perception/emit boundaries). An \emph{agent} is a
\emph{separate} policy (an LLM planner or a fixed rule) that picks actions
toward a goal; it is a \emph{user} of a world, not a part of it, so one world serves
any agent and you can swap the planner without re-synthesizing dynamics. A
\emph{simulation} closes the loop: it puts agents in a world and repeatedly
observes, picks an action, and steps the world, producing a trajectory scored by
objectives. The textbook definition puts the world model \emph{inside} the agent
(the simulator it plans with) and sets it apart from the true environment; the
gap between the two (model bias, sim-to-real error) is exactly what limits learned
world models. A verified code world model does not so much remove that gap as
\emph{relocate} it: once the dynamics are declared symbolically and the accepted
program reproduces them exactly, the model an agent plans through and the (symbolic)
environment it acts in can be the \emph{same} verified transition, so the dynamics add
\emph{no} rollout
error---any error that remains lives entirely in the \emph{specification} and in
\emph{perception} (the symbol-grounding boundary; see the companion framework paper~\cite{openworld_framework}),
not in the transition. This is why the same \texttt{World} serves, unchanged,
both as the ground-truth environment and as the model an agent plans
through in our planning experiments; it is also why this paper is scoped to
\emph{specifiable} domains.

This Code World Model bargain (programs instead of weights) buys verifiability,
compute-cheap rollouts,
and out-of-distribution generalization by construction~\cite{worldcoder2024,
gifmcts2024, poeworld2025, cwmgames2025}. Yet CWM research has stayed a set
of one-off systems aimed at particular benchmarks---built for evaluation or
post-training, rather than giving agents a reusable framework to use them during
inference. No general,
lightweight framework has let a practitioner declare a world, have a
local model write and verify its dynamics, steer behavior through declared
value weights, and optimize the whole setup against a goal---the
workflow that made supervised learning ubiquitous.

\openworld{} is that framework, and this paper benchmarks the paradigm it
puts into practice against the learned-dynamics alternative. The framework
adds four mechanisms: (i) a \emph{plan--generate--verify relay} in
which a generator LLM writes transition code that must pass syntactic
checks, sandboxed smoke-runs, and declared invariants, and optionally a
second-model critic, before we accept it; (ii) \emph{open value
specification}---objectives are scoring functions weighted by tunable dials,
so the moral configuration is an artifact you can edit rather than a frozen weight,
which speaks directly to the specification trap~\cite{spectrap2025}; (iii) an
\emph{automated tuner} that searches world design, policy knobs, and moral
dials together (random, local-refinement, and Optuna TPE~\cite{optuna2019}
strategies); and (iv) \emph{agents-as-a-judge}~\cite{agentasjudge2024,
llmjudge2023}: an LLM judge that picks among candidate behaviors and grades
trajectories against written rubrics.

A survey of the world-model landscape shows distinct strategy families with
a recurring tension---fidelity and scale on one side; verifiability, sample
efficiency, and steerability on the other (a positioning table against the
representative strategies appears in the companion framework paper).

\paragraph{Explicit vs.\ implicit world models, and world models as evaluators.}
Two strands of recent work sharpen what a verified, explicit world model buys. The first
asks whether a trained model even holds a coherent world model: implicit world models
read out of a generative model's predictions are often incoherent~\cite{vafa2024evaluating},
and whether a network represents one inside at all is contested~\cite{li2025what}---a
question \openworld{} steps around, because the model is the human-readable transition program,
so coherence is a property you check, not one you infer. Video-generation world models
likewise miss the governing physical law even at scale~\cite{kang2024how}; \openworld{}
instead declares and verifies the law, trading learned breadth for exactness where the law is
known. The second strand uses world models as policy-evaluation
environments~\cite{quevedo2025worldgym}; because those environments are learned, a policy's
score picks up the simulator's error, whereas an \openworld{} evaluation world runs
exactly, so model error does not confound the score.

\paragraph{Contributions.} Three, all around the world-time-compute thesis:
\begin{enumerate}
  \item \textbf{World-time compute on real domains, with cross-world transfer.}
  A verified world model is \emph{training signal}: fine-tuning an LLM on trajectories run
  through many world models of a domain raises its generalization to held-out worlds it
  never trained on. The new cross-world axis---train on a family, generalize to worlds you
  have not seen---holds on a real, human-authored domain (List Functions, E84): one adapter trained on
  \CrossWorldNTrain{} disjoint worlds reaches \CrossWorldMean{} on \CrossWorldNEval{} held-out
  worlds versus \CrossWorldCorrupt{} for a corrupted-label control---a \mbox{+\CrossWorldGap}-point
  exactness-gated lift (CI \CrossWorldGapCI{}, \CrossWorldNSeeds{} seeds). On synthesized families
  the gain is largest where capability is scarcest ($+\WTSmallGain$ at \WTSmallName, within noise
  by \WTLargeName{} as the task saturates). A corrupted-label control shows that exact labels,
  not task variety alone, are the lever---which is what sets this apart from domain randomization.
  \item \textbf{Why it must be verified code.} Running a world is safe only if the
  worlds do not lie. On three ground-truth worlds, verified synthesized dynamics are exact on
  \MultiCodeExact{} \RolloutSteps-step rollouts (95\% CI \MultiCodeCI) and answer $10\times$
  out-of-distribution probes at \CodeOODRate{} from zero transitions, while the LLM
  next-state proxy completes \MultiLLMExact{} and a 10{,}000-transition MLP scores \MLPOODTenK{}
  out of distribution. We \emph{measure} verification, we do not assume it (each gate buys $+0.24$ probe
  accuracy over blind acceptance), and it certifies executable, invariant-respecting code, not
  rule-correctness (on a weak 3B generator the gate accepts every program yet none is
  branch-exact). Handed the dynamics, planning through the verified model beats planning through a
  hallucinating one and solves $0$-shot what trained DreamerV3 / V-JEPA-2 need $10^4$--$10^5$
  steps to learn~\cite{dreamerv3} (a scope property, not a contest; \S\ref{sec:bakeoff}).
  \item \textbf{The substrate, and full reproducibility.} \openworld, a zero-dependency framework,
  turns authoring and serving the many verified worlds world-time compute needs into a
  minutes-scale activity; the framework, its \ProtoNumWorlds-world prototyping benchmark, and the
  broader world zoo are the companion paper~\cite{openworld_framework}. All \NumExperiments{}
  experiments are seed-fixed, use paired tests, and regenerate from one public repository.
\end{enumerate}

\paragraph{Positioning in the field.} \openworld{} builds on the \emph{programmatic} code-world-model
program~\cite{worldcoder2024,gifmcts2024,poeworld2025,cwmgames2025}---LLM-synthesized, verified
code dynamics, as distinct from \emph{neural} code world models that learn dynamics in
weights~\cite{metacwm2025} and from LLM synthesis of symbolic world models via test-time
scaling~\cite{ivml2025}---but its contribution is what that machinery makes possible: world-time compute,
which turns many verified worlds into exactly-labeled training signal. Set against the
\emph{learned} world-model families---latent-dynamics models (Dreamer~\cite{dreamerv3},
MuZero~\cite{muzero}), tokenized and diffusion video models (IRIS~\cite{iris2023},
DIAMOND~\cite{diamond2024}, Genie~\cite{genie2024}, Sora~\cite{sora2024}), and
joint-embedding predictors (V-JEPA~\cite{vjepa2025})---it trades perceptual breadth for an
auditable, exactly-executable symbolic transition with no compounding rollout error. Its closest relatives
are the \emph{self-training} methods that retrain a model on its own verifier-filtered outputs
(STaR~\cite{star2022}, ReST~\cite{rest2023,restem2024}, expert iteration~\cite{expertiteration2017},
rejection-sampling fine-tuning~\cite{rft2023}, verifiable rewards~\cite{rlvr2024}, and test-time
reinforcement learning on unlabeled data via majority-vote pseudo-rewards~\cite{ttrl2025}); we differ in
what does the verifying (a full world model with dynamics, so the unit is a verified trajectory, not
a checked answer) and in the \emph{generalization axis} (held-out worlds in a family). It also
differs from domain randomization and procedural task generation~\cite{tobin2017dr,cobbe2020procgen,openended2021xland}---and
from foundation-model-generated code environments for open-ended learning~\cite{omniepic2024}---where
the experience comes from a single (often hand-coded or learned) simulator, or from generated
environments picked for how interesting they are rather than checked for label-exactness. Here every
world is synthesized and
verified on its own, and the open question we pose is whether that label exactness, not task
variety, drives the held-out gains. The labels are
exact only \emph{relative to the synthesized world model}---``verified'' means the transition
respects the declared invariants and passes the probe set, not that it provably equals the
domain's true dynamics (which we may only observe in part). So a world whose declared rules are subtly wrong is
a confidently-wrong teacher; the corrupted-label ablation (\S\ref{sec:world-time-scale})
measures exactly that failure mode. The supporting machinery extends the open-specification
stance of the value-alignment literature~\cite{spectrap2025,mas2025} and judge-based
behavior selection~\cite{agentasjudge2024}. We benchmark on consumer hardware with 7B/3B
local models, the regime where learned world models are least accessible and training-free
approaches matter most.

\section{Problem Setting}
We aim at symbolic-state environments. In these worlds the state is a
JSON-serializable structure, the action set is discrete, and the dynamics
follow rules you can declare. This covers operational simulations (triage,
negotiation, portfolios, sprints) and program repair, but it does not cover
pixel-native domains; we come back to this boundary in the limitations. The
practitioner declares a world---its state schema, its actions, and its rules
in plain language. The framework must then build a simulator that is faithful,
fast, and auditable, plus the tools to steer and optimize behavior inside it.
What can go wrong here is \emph{model error itself}: made-up transitions,
silently wrong code, badly specified rewards, and value drift.

\paragraph{A world model, formally.} An \openworld{} world model is a tuple
\[
  W \;=\; \bigl(\mathcal{S},\ \mathcal{A},\ s_0,\ \tau,\ \{(g_i, w_i)\}_i,\ \mathcal{I}\bigr),
\]
with symbolic states $\mathcal{S}$ (JSON-serializable maps), a discrete action set
$\mathcal{A}$, an initial state $s_0$, a \emph{deterministic} transition program
$\tau : \mathcal{S} \times \mathcal{A} \to \mathcal{S}$ (illegal actions do nothing, so $\tau$
is total; stochastic worlds thread a seed through, so rollouts stay replayable), weighted
objectives $\sum_i w_i o_i$ that score transitions, and invariants $\mathcal{I}$ that must
hold on every reachable state. The practitioner declares everything but $\tau$---the schema of
$s_0$, the actions, the natural-language rules $R$, the objectives with their \emph{dials},
and the invariants---and the framework must \emph{produce} $\tau$. The
framework's full \texttt{World} anatomy is wider---\emph{perception}\,$\to$\,state\,$\to$\,actions\,$\to$\,transition\,$\to$\,reward
(the companion framework paper~\cite{openworld_framework}); here we work with
its dynamics-and-scoring core and treat perception, the map from a raw
observation to a symbolic state, as a separate concern. Making perception a
first-class part---grounding a world directly from raw observation such as
pixels---is the focus of a companion ARC-AGI-3 paper~\cite{openworld_arc3} (under review).

World-time compute does not roll the transition forward. Instead the agent takes a world
model \emph{it built and verified} as ground truth, and it trains on the exactly-labeled
examples that world hands it: an input $o$ paired with the answer $y$ the world fixes---the
disease a generative world drew (E74), the output a real task's hidden rule determines (List
Functions, ARC), or the next state a verified program computes (program repair). The label is
exact because the agent reads it off its own verified world rather than guessing it---exact
\emph{relative to} that world. The question this paper sets out to measure is how far an agent
that builds its own worlds and learns from them can then handle wide-ranging problems.

\openworld{} produces $\tau$ by \emph{synthesis under verification}, not by learning. An LLM
proposes a program $\hat\tau$ from $R$. We accept it only if it parses with the required
signature, runs in a sandbox that keeps every invariant on a probe set, and, optionally, gets
past a second model's critique. This certifies that $\hat\tau$ is invariant-consistent, not
that it equals the true dynamics $\tau^\star$. But once we accept it, $\hat\tau$ is fixed code:
wherever it matches $\tau^\star$ on the reachable set, a $t$-step rollout is exact at every
depth, bit-for-bit. A transition predicted step by step with single-step error rate $\epsilon$
instead stays exact for only about $(1-\epsilon)^t$ of a $t$-step rollout, decaying with
depth---the compounding error that a fixed program avoids by construction.

\section{Experimental Setup}
\paragraph{Worlds with ground truth.} Three deterministic oracle worlds drive
every dynamics comparison: \emph{sprint} (backlog/debt/bug
dynamics with an integer-division interaction), \emph{orchard} (a resource pool
with per-agent tallies), and \emph{triage} (queues, a clock, and
deterioration events). The oracles are hand-written
\texttt{FunctionTransition}s. We compare synthesized and learned-style engines
against them exactly, including on a fixed probe suite that
exercises the clamping, empty-queue, and interaction branches.

\paragraph{The coding world.} Program repair is the flagship use case: it is
well-known, objectively scored, and symbolic by nature~\cite{humaneval2021}.
Each of \NumTasks{} tasks is a buggy Python function plus a hidden test suite
that spans classic defect archetypes (off-by-one ranges and binary-search
bounds, inverted comparisons, missing normalization, wrong base cases,
unsorted medians, order-destroying dedup, early imbalance, whitespace
splitting, accumulator resets, integer-division truncation, overlapping
counts, dropped final runs, and degenerate-input handling). The world's
transition is test execution: a submitted patch runs bit-exactly in a
sandbox with a wall-clock alarm (a looping patch fails instead of hanging),
and failing-test feedback enters the state for the next attempt.

\paragraph{Benchmark-scale repair (E28--E29).} Beyond the single-function
benchmark, two SWE-bench-style suites test repair at module scale. Each
instance carries a natural-language issue report (symptoms and a repro, never
the fix), a buggy module (functions or a stateful class), two hidden suites
(\texttt{fail\_to\_pass} exercising the defect, \texttt{pass\_to\_pass}
guarding against regressions), and an explicit world specification whose
\texttt{submit\_patch} dynamics run both suites bit-exactly. Solving requires
zero failures in both, so a fix that breaks a passing test does not
count. The \emph{atomic} suite (E28, $n=\SweAtomicN$) plants one
issue-described defect per instance; the \emph{staged} suite (E29,
$n=\SweStagedN$) plants a second, hidden defect that shows up as a new
failing test only once the issue-visible defect is fixed. Both run a paired
ablation across the \texttt{qwen2.5} ladder (1.5B/3B/7B): the same model
\emph{single-shot} (one completion from issue + module) versus \emph{in-world}
(up to \SweBudget{} \texttt{submit\_patch} steps; the first prompt is
identical to single-shot, and exact failing-test feedback enters from the
second attempt onward). An expanded 20-instance atomic suite with extra
regression traps ships in the repository
(\texttt{datasets/openworld-repairbench/}).

\paragraph{Composition, nesting, and changing rules (E30--E32).} Three
experiments exercise \texttt{CompositeWorld}, a wrapper whose state nests its
children's states under namespace keys. Children run \emph{unmodified}: the
composite slices out a child's namespace, steps the child's own transition,
and writes it back. Coupling is explicit. It runs sideways through
\texttt{Bridge}s (an ordinary transition over the two-slot dict of the coupled
children's states, so the same synthesis-and-verification pipeline applies to
bridges unchanged), and upward through \texttt{Aggregator}s (derived from the
leaves, never simulated); agents travel between children over \texttt{Route}s
whose crossing effects (tolls, vetoes) are themselves verifiable transitions.
\texttt{PhasedTransition} handles rules that change over time: ordered
(trigger, transition) phases that advance in sequence and cannot reverse,
with every phase built and verified before the run. E30 holds one system
fixed---\CompRules{} internal rules plus four ring couplings---and varies
only how we pose the synthesis: one monolithic prompt over a flat namespaced
state versus eight small prompts (four \CompChildRules-rule sectors, four
one-rule bridges), each verified on its own and then assembled into a
\texttt{CompositeWorld}. Both conditions score on the same ground-truth probe
suite (7B generator, three replicates each). E31 replays a \NestSteps-step
script (leaf actions, ticks, and two travel attempts across a toll route) on
a three-level composite against an independent flat-dict oracle that imports
nothing from the composition machinery, checking leaf exactness, aggregator
consistency, money conservation, and agent-registry agreement at every step.
E32 declares a market whose policy changes at a fixed step mid-rollout, and it
compares phased synthesis (each phase from its own rule text alone),
monolithic synthesis from the combined rules-with-change text, and the LLM
next-state proxy, on closed-loop rollouts that cross the boundary.

\paragraph{Models and hardware.} All experiments use local models served by
Ollama on an Apple-silicon laptop: \texttt{qwen2.5:7b} (generator, judge,
critic in E3), \texttt{qwen2.5:3b} (small generator in E2/E3),
\texttt{qwen2.5:1.5b} (repair agent in E5/E6/E13/E17), and---for
cross-family replication (E16)---\texttt{llama3.1:8b} (Meta) and
\texttt{gemma2:9b} (Google). The trained baselines in E12/E19 are plain
numpy models. We use no cloud APIs, no fine-tuning, and no training compute
beyond the seconds-scale fitting of the baselines.

\paragraph{Baselines.} We represent the learned-dynamics family in two ways.
(i) \texttt{LLMTransition}: the same backbone predicting each next state
directly from the same description and rules---it shares the symbolic engine's
knowledge but has the per-step stochastic prediction structure of
learned models. (ii) Trained models (E12): a two-hidden-layer
MLP and a 1-nearest-neighbor memorizer, each trained on $K \in \{100, 1000,
10000\}$ environment transitions collected by a random policy---real learned
dynamics with a measured sample-efficiency curve. We say plainly that
neither is a trained Dreamer; reimplementing the
pixel-native family at laptop scale is infeasible, and that is itself part of the comparison
(DreamerV3-class systems require millions of environment
steps~\cite{dreamerv3}).

\paragraph{Metrics and statistics.} Rate metrics carry 95\% Wilson confidence
intervals. Dynamics fidelity uses exact-state match (every field equal),
the first-divergence step, and final-state $L^1$ error. Program repair reports
pass@1 and pass@budget (4 attempts); the controlled selection comparison
(E13) is a paired design on shared candidate sets with an exact two-sided
McNemar test. Judge alignment uses Spearman rank correlation with permutation
p-values (10{,}000 shuffles). We fix the seeds in each script; every number in
this paper regenerates via \texttt{python3 scripts/make\_paper\_assets.py}.

\paragraph{Calibration, pilots, and internal review.} We report the design
changes we made after pilots rather than hide them: (a) the verifier ablation (E3)
uses the 3B generator at high temperature because the 7B generator's first
attempts passed every gate, leaving nothing for the conditions to tell apart;
(b) the repair agent in E5/E6/E13 is the 1.5B model because both the 7B and
3B agents saturated the original benchmark (pass@1 $=100\%$)---with
failing-test feedback in the state, these archetypes are easy for capable
models. The capability ladder (1.5B: 70\%, 3B and 7B: 100\% on the original
suite) is itself a finding. All LLM-dependent numbers are
specific to the stated Ollama snapshots and Q4\_K\_M quantization.

\section{Results}\label{sec:results}

\subsection{Head-to-head: symbolic vs.\ learned-style dynamics}
This section measures \emph{dynamics fidelity}---how well we predict a world's
next state from the current state and an action---not the program-repair task of
the coding world. The practitioner writes down each world's rules. The question
is how to turn those written rules into a faithful simulator. The symbolic engine
compiles the rules to a verified \texttt{transition} program once, then runs it.
The learned-style engine predicts each next state directly. The sprint world's
\texttt{bugs\,{+}{=}\,debt//4} coupling is an arbitrary \emph{declared} rule, not
a natural law to be guessed. The point is not that a model should see it coming.
The point is that once you declare it, compiling it to code runs it exactly,
while predicting it step by step piles up error.

Table~\ref{tab:main} and Figure~\ref{fig:hero} show the main comparison
(E1, E4, E10), and Table~\ref{tab:fidelity} runs the same protocol across
all three instrumented worlds, with eight action scripts each (E11). The
verified synthesized engines are exact on \MultiCodeExact{} rollouts (95\%
CI \MultiCodeCI), against \MultiLLMExact{} (CI \MultiLLMCI) for per-step LLM
prediction. And because planning runs Python instead of a forward
pass, they roll out at \CodeStepsPerSec{} steps/s against the LLM engine's
\LLMStepsPerSec{} steps/s, after a one-time synthesis cost of a few seconds.

\begin{table}[htb]\centering\small
\begin{tabular}{lcccc}
\toprule
Engine & First divergence (step) & Final L1 error & OOD exact (10$\times$) & Steps/s \\
\midrule
\textbf{Symbolic (verified code, ours)} & 21.0 & 0.0 & 100\% & 14,997 \\
LLM next-state (proxy) & 2.3 & 10.3 & 40\% & 0 \\
\bottomrule
\end{tabular}

\caption{Head-to-head engine comparison on the sprint world (E1, E4, E10):
mean first-divergence step and final $L^1$ error over three
\RolloutSteps-step rollouts against a ground-truth oracle; exact transition
accuracy on ten $10\times$ out-of-distribution probes; and rollout
throughput. A first divergence of \CodeFirstDivergence{} means no divergence
occurred within the rollout.}
\label{tab:main}
\end{table}

\begin{table}[htb]\centering\small
\begin{tabular}{lccc}
\toprule
World & Code: exact rollouts & LLM: exact rollouts & LLM first divergence \\
\midrule
orchard & 8/8 & 0/8 & 11.2 \\
sprint & 8/8 & 0/8 & 1.4 \\
triage & 8/8 & 0/8 & 2.0 \\
\midrule
\textbf{Total} & \textbf{24/24} [0.86, 1.00] & 0/24 [0.00, 0.14] & --- \\
\bottomrule
\end{tabular}

\caption{Multi-world replication (E11): exact \RolloutSteps-step rollouts per
world, eight scripts each, dynamics synthesized once per world by the 7B
generator. Totals carry 95\% Wilson CIs.}
\label{tab:fidelity}
\end{table}

\subsection{The central result}\label{sec:central-result}
\emph{Verified code synthesis eliminates compounding rollout error at zero
training cost.} The reason is structural, not statistical. A learned (or
per-step predicted) transition is sampled fresh each step, so with per-step error
$\epsilon$ the probability that a $t$-step rollout has \emph{diverged} grows as
$1-(1-\epsilon)^{t}$ (one minus the exactness probability $(1-\epsilon)^t$): with the measured
single-transition error of the LLM engine (\LLMProbeErrorRate{} of probes wrong),
divergence by step \LLMFirstDivergence{} is just what the arithmetic
predicts, and twenty-step plans are fiction (final $L^1$ error
\LLMFinalLOne). The symbolic engine pays its whole error budget once,
at synthesis time, where you can audit it: if the accepted program is right,
every rollout of any depth is right, bit-exactly. This is the CWM
hypothesis~\cite{worldcoder2024,cwmgames2025} made measurable on consumer
hardware.

The single-number divergence \LLMFirstDivergence{} comes from one world (sprint,
$n{=}3$ rollouts, one seed), and it is not representative: across the three
worlds (E11, \MultiWorldRollouts{} rollouts) the LLM proxy's first divergence
ranges from step ${\approx}1.4$ (sprint) and $2.0$ (triage) to ${\approx}11$ (orchard),
because compounding depends on the per-step error rate, and that rate varies by world.
The claim that holds regardless of world is the exact-match contrast, not the mean
divergence step: code engines finish \MultiCodeExact{} of \MultiWorldRollouts{}
twenty-step rollouts bit-exactly (95\% CI \MultiCodeCI) versus \MultiLLMExact{}
for the LLM proxy. We treat the divergence figure as an illustration of the
mechanism and the exact-match rate as the result.

\subsection{World-time compute: traversing world models to generalize from fewer examples}\label{sec:world-time}
The results so far buy a \emph{correct} world model cheaply. A verified world model is also a
\emph{generator}: it makes unlimited, exactly-labeled experience and shows its own
rules, so an agent can \emph{traverse} many worlds of a domain and distill the shared skill.
We call spending compute this way---fine-tuning a model on trajectories drawn from traversed,
verified world models rather than on scarce real examples---\textbf{world-time compute}, a
training-time analogue of test-time compute~\cite{snell2024testtime,openai2024o1,deepseekr1}
(and, when spent per world at inference, of test-time training (TTT)~\cite{akyurek2024ttt}).

\begin{figure}[t]\centering
\includegraphics[width=0.98\linewidth]{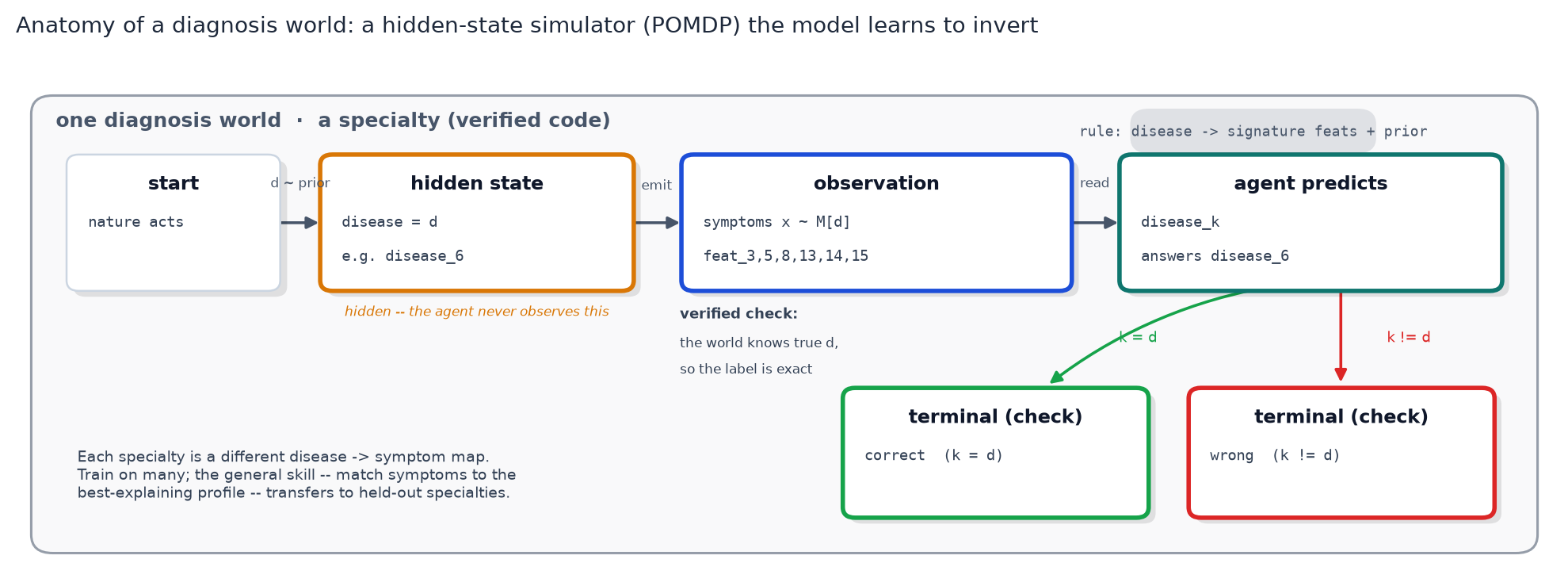}
\caption{\textbf{Anatomy of one diagnosis world (E74--E79).} Each specialty is a small
hidden-state simulator: nature draws a hidden disease from a prior, the world emits
symptoms from that disease's signature features (plus background noise), and the agent
must invert the observation to name the disease. The agent treats this world---which
it built and verified---as ground truth, so the disease the world drew is a known, exact label
(relative to that world): a traversed world distils a trusted rule, not a hallucinated one. Each specialty is a different symptom\,$\to$\,disease
map; training on many and testing on held-out specialties forces the general skill
(match symptoms to the best-explaining profile) rather than memorisation.}
\label{fig:diagnosis-world}
\end{figure}

We test it on a \emph{family} of diagnosis worlds (E74): one partially observable Markov decision
process (POMDP) template (a hidden
disease; order tests to reveal symptoms; commit a diagnosis) instantiated as
\WTNumTrainWorlds{}${+}$\WTNumTestWorlds{} specialties that share a goal and action grammar
but differ in their symptom\,$\to$\,disease structure. We hold out whole specialties---a
world-level train/test split, as in supervised learning: fine-tune a model on traversed
train specialties, then measure diagnostic accuracy on \WTNumTestWorlds{}
unseen specialties (\WTNumCases{} cases). Two brackets frame the result: a prior-only floor (\WTFloor)
and an oracle handed the rules (\WTOracle). Figure~\ref{fig:diagnosis-world} walks through one
such world---a hidden-state POMDP the model learns to invert. To be exact about what
transfers: each test prompt states that specialty's profiles (its rules) in context, and
the base baseline sees the same prompt---so the lift is not the fine-tuned
model reading a table the baseline cannot (both see the rules). The lift is fine-tuning teaching the
model to apply the given rules in a way that carries to symptom\,$\to$\,disease maps it
never trained on. This is the in-context regime of test-time training, one rung more general
(across rule-sets, not within one); internalizing fully \emph{latent} dynamics is harder and we
do not claim it here.

Two findings (Figure~\ref{fig:hero}B). \emph{(i) Generalization, not memorization:}
world-time compute lifts held-out accuracy at every model size---from \WTSmallBase{} to
\WTSmallFT{} at \WTSmallName{}, and from \WTLargeBase{} to \WTLargeFT{} at \WTLargeName{}---%
moving the model toward the rules-given oracle on specialties it never trained on.
\emph{(ii) It is a small-model multiplier:} the gain is largest where capability is
scarcest ($+\WTSmallGain$ points at \WTSmallName, bootstrap CI over held-out specialties
well clear of zero) and \emph{shrinks} as larger models approach the rules-given ceiling---%
by \WTLargeName{} the lift is only $+\WTLargeGain$ points and its CI includes zero, because
the task is nearly saturated (oracle \WTOracle). Does the gain come back once capable
models are given headroom---a deliberately harder world family? That is a clean falsifiable
follow-up we are running. An offline analogue confirms the distilled structure is a
sample-efficiency win either way: a learner that has seen the family diagnoses a held-out
specialty from fewer cases than one learning from scratch (\WTMetaFewShot{} vs
\WTScratchFewShot{} at a single case per disease).

The family is what makes this work. Fine-tuning on \emph{heterogeneous}, unrelated
worlds that share no task structure yields no such transfer (E71, E73): there is no shared
skill to distill, so a policy expert on one world does no better than chance on another.
World-time compute pays off exactly when the traversed worlds form a domain---then more
traversal becomes a lever on generalization, complementary to train-time data and test-time
search. (Scope: the model-size sweep mixes precisions---0.5B--7B are bf16 LoRA, the larger
sizes 4-bit QLoRA---and the per-number-of-worlds axis saturates quickly, so we claim a
compute \emph{lever}, not a clean world-count scaling law.)

\subsection{Scaling the lever, and a coding-world transfer test}\label{sec:world-time-scale}
Two follow-ups sharpen the principle. Does it scale with the number of worlds? And
does it hold on a different use case, one authored as real verified-code worlds?

\begin{figure}[t]\centering
\includegraphics[width=0.82\linewidth]{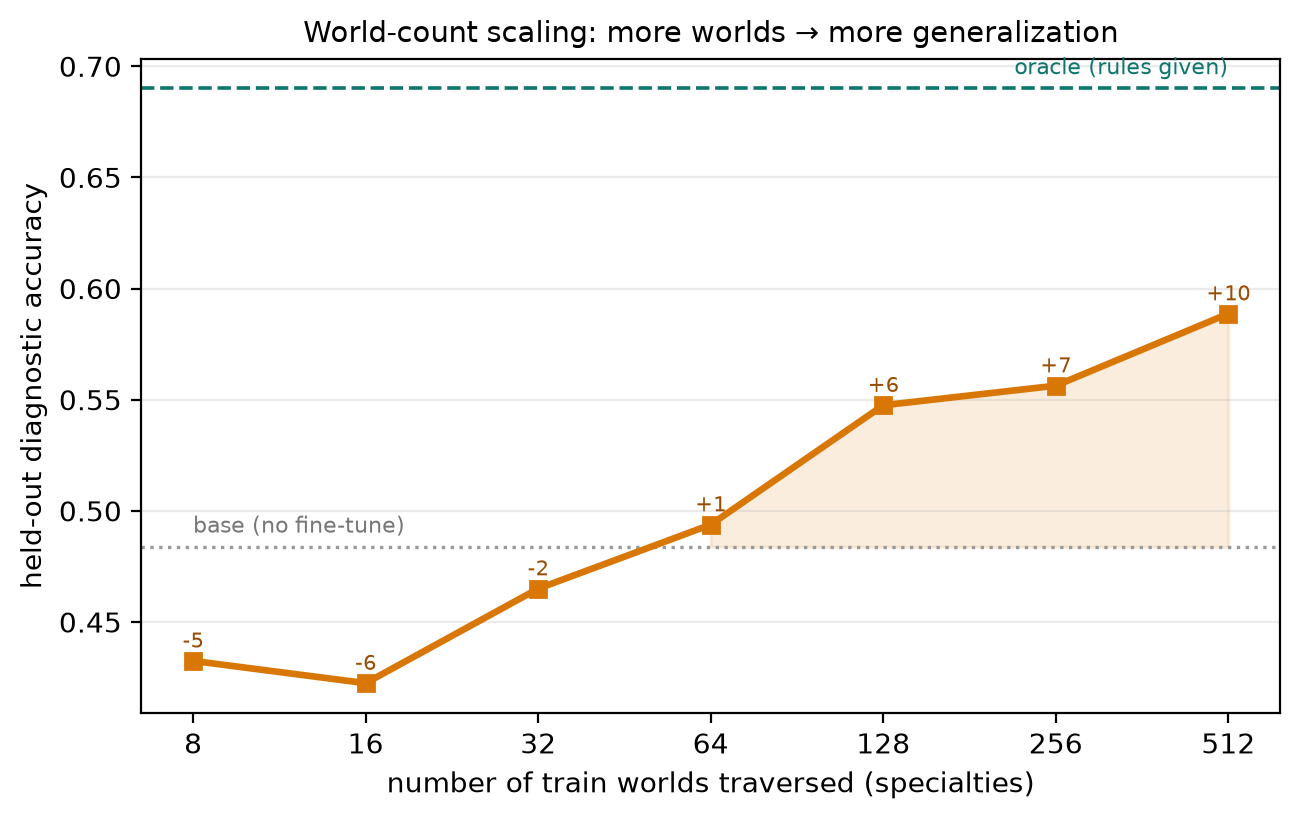}
\caption{\textbf{World-count scaling (E76).} Held-out diagnostic accuracy of a 7B model
vs the number of train specialties it traversed (hard family; test specialties fixed).
Labels are the gain over the no-fine-tune base. Few worlds hurt; the gain rises
monotonically and has not plateaued at \WCMaxWorlds{}, climbing toward the rules-given
oracle. More worlds $\to$ more generalization.}
\label{fig:world-count}
\end{figure}

\paragraph{World-count scaling (E76).} We fix the model (7B) and the hard family and vary
only the number of train specialties traversed, holding the test specialties fixed
(Figure~\ref{fig:world-count}). Too few worlds \emph{hurt}: the fine-tune overfits a narrow
slice, below the \WCBase{} base. After an initial dip the gain rises monotonically and is \emph{still
climbing at \WCMaxWorlds{} worlds} (\WCMaxAcc, $+\WCMaxGain$ points), heading toward the
\WCOracle{} oracle---no plateau. So world-time compute is a genuine scaling lever in the
number of worlds, when the task has headroom. The offline empirical-Bayes prior saturated at
${\approx}2$ specialties precisely because it has none of the world's \emph{skill} left to keep
learning.

\begin{figure}[t]\centering
\includegraphics[width=0.7\linewidth]{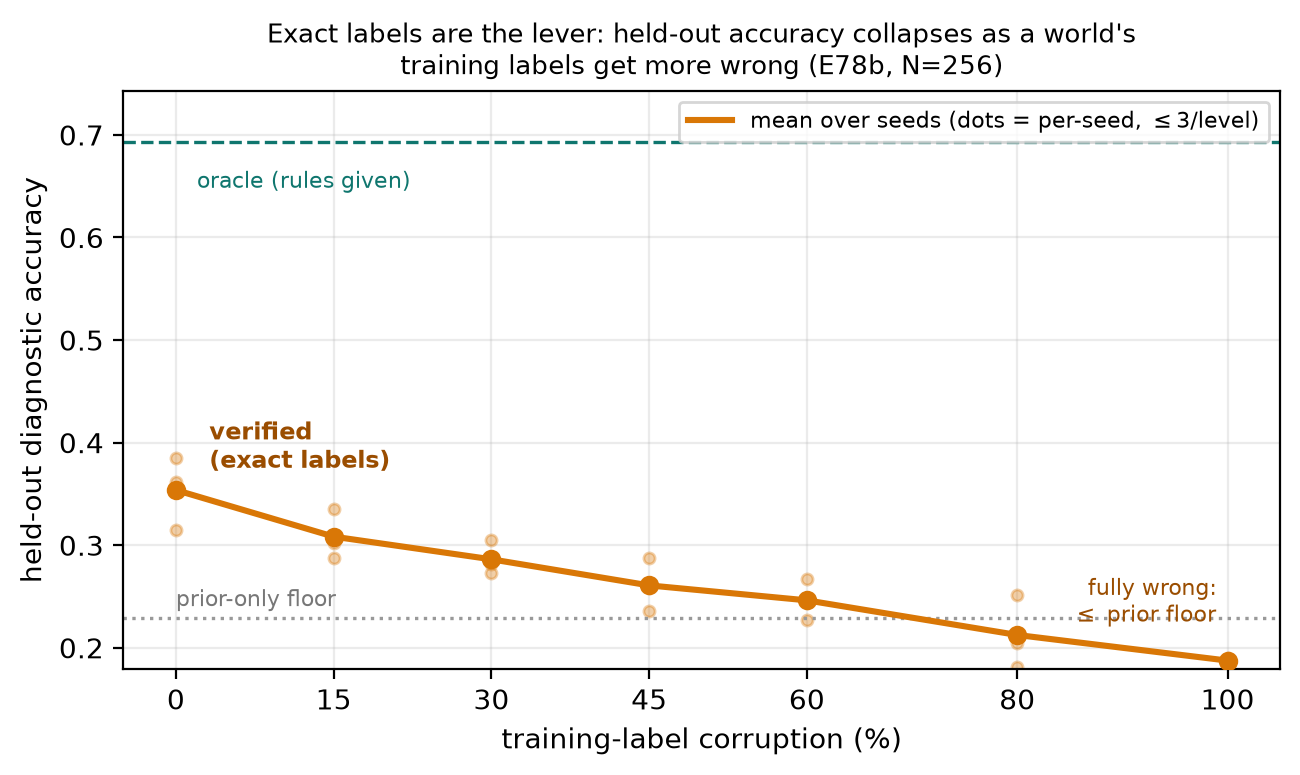}
\caption{\textbf{Exact labels are the lever (E78b).} Held-out diagnostic accuracy after
fine-tuning on \AblWorlds{} worlds whose training labels are corrupted at rate $p$
(same worlds, patients, and prompts throughout; the test labels are always exact). Verified
worlds ($p{=}0$) sit at the top; accuracy falls monotonically to the no-fine-tune base as the
world becomes fully wrong ($p{=}1$). What world-time compute needs is label exactness, not task variety. Dots are per-seed values (\AblSeeds{} seeds/level).}
\label{fig:noise-ablation}
\end{figure}

\paragraph{Exact labels are the lever (E78b).} Scaling shows that more worlds help; this
ablation shows why the worlds must be verified. We hold the world count fixed
(\AblWorlds{} train specialties) and keep the patients, prompts, and example count identical, and
vary only the training labels: a fraction $p$ of each world's diagnoses is replaced by
a confusable disease drawn from the posterior over the others---the structured error a
non-verified or learned world model makes when its rules are slightly wrong. The held-out
test labels stay exact. Held-out accuracy falls monotonically as corruption rises, from
\AblVerified{} at $p{=}0$ (verified) down to \AblNoisyFull{} at $p{=}1$---a fully-wrong world
gives \emph{no} generalization gain---in fact at or below the prior-only floor (\AblFloorHard)---with a smooth penalty in
between (Figure~\ref{fig:noise-ablation}). So what drives the effect is label \emph{exactness}, not task variety---the property that sets world-time compute apart from domain
randomization, and the reason the worlds must be verified code (\AblSeeds{} seeds per
level; per-seed values shown in Figure~\ref{fig:noise-ablation}).

\paragraph{A coding world family (E77).} Coding is the cleanest verified-code world we have:
the transition is execution and the oracle is the tests passing. We had a model author
\CodeNumWorlds{} function-implementation tasks, each admitted only after its reference
solution passes its own sandboxed \texttt{pytest} oracle, and we fine-tuned on the traversed
worlds (Table~\ref{tab:coding}). \emph{In-domain}, world-time compute lifts best-of-$k$ at
every model size (7B pass@5 \CodeSynBaseFive$\to$\CodeSynFtFive). On HumanEval---a
real, out-of-distribution benchmark---it hurts greedy pass@1
(7B \CodeHeBaseOne$\to$\CodeHeFtOne; the synthetic fine-tune narrows the single best guess)
yet \emph{transfers positively at pass@5} (\CodeHeBaseFive$\to$\CodeHeFtFive) from only
\CodeNumWorlds{} worlds. That fits E76: \CodeNumWorlds{} worlds is below the
count where transfer becomes strong, so the path to real-benchmark gains is to scale the world family, not to abandon the
approach.

\begin{table}[t]\centering\small
\begin{tabular}{lcccc}
\toprule
Model & Synthetic p@1 & Synthetic p@5 & HumanEval p@1 & HumanEval p@5 \\
\midrule
0.5B & 0.17$\to$0.27 & 0.38$\to$0.55 & --$\to$-- & --$\to$-- \\
1.5B & 0.51$\to$0.55 & 0.78$\to$0.84 & --$\to$-- & --$\to$-- \\
3B & 0.78$\to$0.70 & 0.85$\to$0.91 & --$\to$-- & --$\to$-- \\
7B & 0.73$\to$0.80 & 0.84$\to$0.95 & 0.78$\to$0.70 & 0.87$\to$0.91 \\
\bottomrule
\end{tabular}

\caption{\textbf{Coding world family (E77).} Sampled pass@1/pass@5 (n{=}5), base
$\to$ world-time-compute (fine-tuned on \CodeNumWorlds{} verified coding worlds), on
synthetic held-out tasks (all sizes) and HumanEval (7B). World-time compute helps in-domain
best-of-$k$ at every size; on HumanEval it hurts greedy pass@1 but transfers positively at
pass@5.}
\label{tab:coding}
\end{table}

\subsection{Real data: world-time compute on ARC-AGI (E80)}\label{sec:world-time-arc}
\begin{figure}[t]\centering
\includegraphics[width=\linewidth]{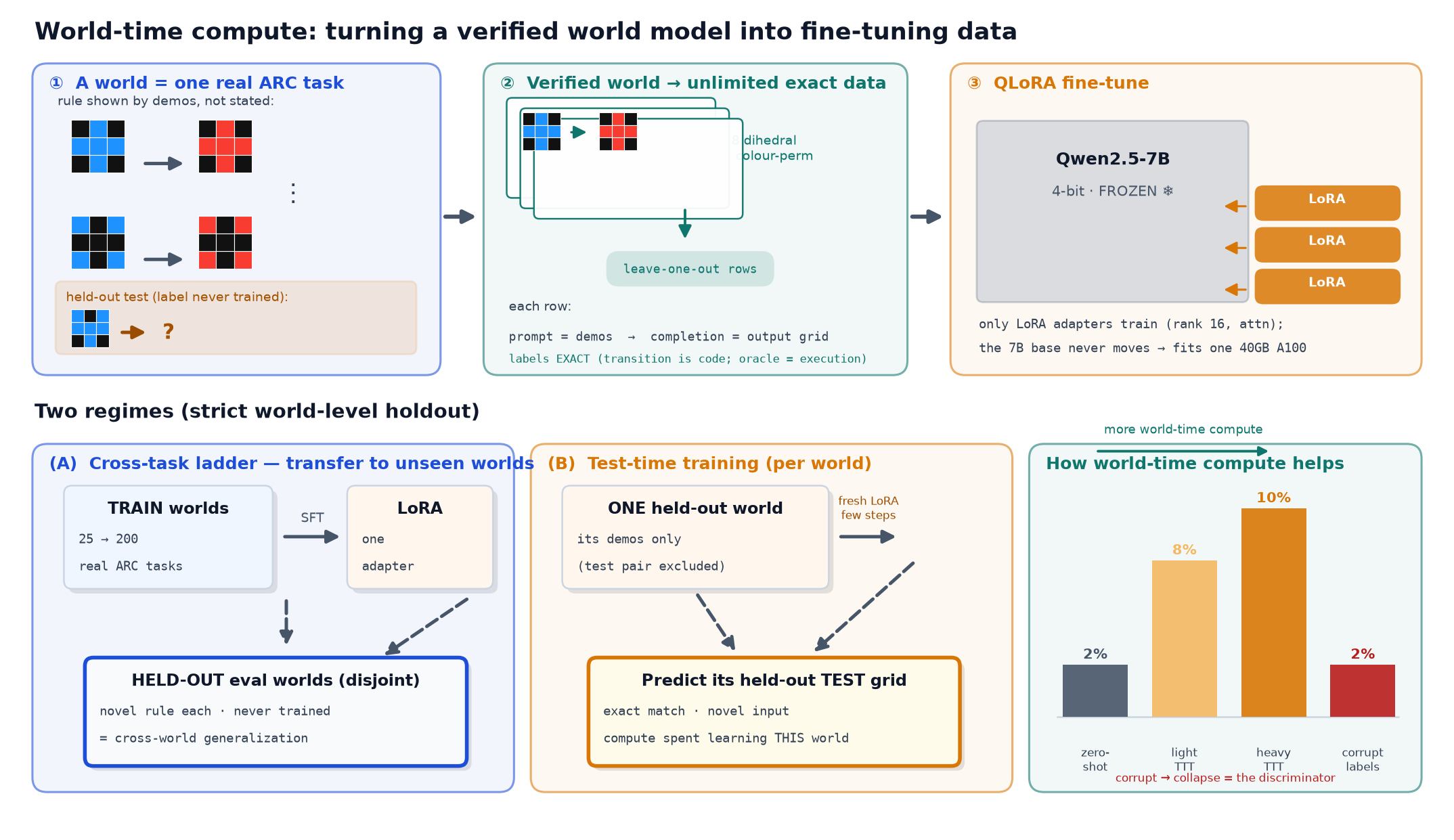}
\caption{\textbf{How world-time compute fine-tunes on a verified world model (E80, ARC-AGI).}
\emph{Top:} each real ARC task is a world whose rule is shown by demonstrations, not stated;
rule-preserving augmentation (dihedral $\times$ color) turns it into unlimited
\emph{exactly}-labeled rows (the transition is code, so the oracle is execution), which
LoRA-SFT a frozen 4-bit Qwen2.5-7B---only the small adapters train. \emph{Bottom:} two
strict world-level holdouts---(A) the cross-task ladder trains on 25--200 worlds and is scored
on \emph{disjoint} held-out worlds; (B) test-time training fits a fresh adapter on a single
held-out world's demos (its test pair excluded) and predicts that test. The payoff bars track
exact-match against world-time compute, with a corrupted-label arm as the discriminator: if the
lift survives label corruption, then the verified labels---not the optimizer---are not what does the work.}
\label{fig:arc-method}
\end{figure}

\begin{figure}[t]\centering
\includegraphics[width=0.96\linewidth]{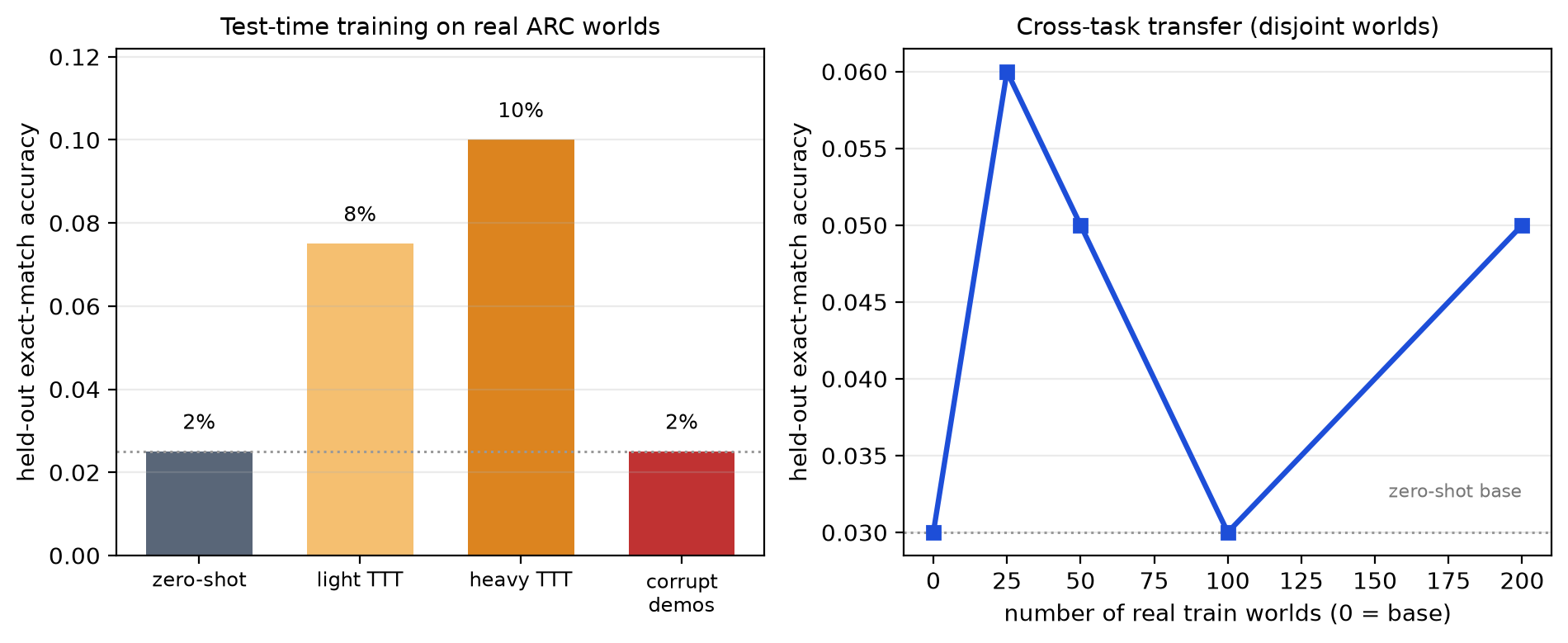}
\caption{\textbf{World-time compute on real ARC-AGI (E80).} \emph{Left:} per-world test-time
training lifts held-out exact-match from zero-shot through light to heavy test-time compute; a
corrupted-demonstration arm (wrong labels, same structure) shows verified labels are
needed---the lift vanishes under corruption (\ArcTTTCorrupt). \emph{Right:} cross-task transfer (train on $N$
real worlds, evaluate on disjoint held-out worlds) is weak, as we expect from ARC's per-task
novelty. Single 4-bit Qwen2.5-7B; mechanism, not leaderboard.}
\label{fig:arc}
\end{figure}

So far, every world-time-compute result runs on worlds we wrote ourselves (diagnosis,
coding). That invites the sharpest objection: maybe the gains come from our generator, not from
the world models. So we move the mechanism to a benchmark we did not write---%
\textbf{ARC-AGI}~\cite{chollet2019arc,arcprize2024}, the human-authored
abstraction-and-reasoning corpus (\ArcNumTrainWorlds{} training + \ArcNumEval{} evaluation
tasks; we use ARC-AGI-1, with ARC-AGI-2~\cite{chollet2025arcagi2} a harder successor). It fits
unusually well. Each task is a world---a hidden grid-transformation rule shown only through a few
input\,$\to$\,output demonstrations. So unlike our synthetic worlds, where the rule is
\emph{stated} in the prompt, here the rule is \emph{latent} and the model must induce it. Labels are
exact: a predicted output grid is either right or wrong, with no judge. The train and evaluation
tasks do not overlap, and each has a novel rule, so ``generalize to held-out worlds'' is the
ARC challenge. We grow each task into many world variants with rule-preserving augmentation
(the dihedral group $\times$ color permutations) and run two tests on a 4-bit Qwen2.5-7B.
The per-task test-time-training recipe---fit a fresh LoRA on a task's augmented
demonstrations---follows Aky\"urek et al.~\cite{akyurek2024ttt}, who showed it is the
effective approach on ARC (cf.\ induction/transduction program search~\cite{li2024barc}). Our
contribution is not the recipe. It is to read the recipe as world-time compute on a verified world,
and to add the corrupted-label ablation that pins down whether the exactness of the
labels---not fine-tuning by itself---drives the gain.
Figure~\ref{fig:arc-method} shows the regime end to end---world model
$\to$ augmented, exactly-labeled rows $\to$ QLoRA fine-tune (frozen 4-bit base, trainable
LoRA), with the two strict train/test splits---and Figure~\ref{fig:arc} reports the results.

\emph{(i) Cross-task transfer is weak.} Train one LoRA on $N$ real training worlds and
evaluate it zero-shot on disjoint held-out worlds, and exact-match barely moves
(\ArcLadderBase{} base $\to$ \ArcLadderTop{} at \ArcLadderTopN{} worlds). This is the
boundary we expect: every ARC rule is novel by design, so a single shared policy
has little to carry over. This matches E76/E77, where transfer only gets strong well above
the world counts we can afford here.

\emph{(ii) Test-time training---world-time compute spent \emph{per world}---works.} For each
held-out world we fit a fresh LoRA on that task's own demonstrations (its test pair
strictly held out), then predict its test grid. This is the thesis in its purest form: compute
spent learning a verified world buys generalization inside it. Exact-match climbs from
\ArcZeroShot{} zero-shot to \ArcTTTLight{} (light) and \ArcTTTHeavy{} (heavy)---$+\ArcTTTGain{}$
points as we spend more world-time compute. The decisive mechanism check is the
\textbf{corrupted-demonstration ablation}. Randomize the demo outputs (same grid structure,
wrong labels), and the lift should vanish if the model is exploiting the exact labels rather than
just gaining from fine-tuning. It does---the corrupt arm falls to \ArcTTTCorrupt{}
(Figure~\ref{fig:arc}), at or below the \ArcZeroShot{} zero-shot floor. So the
lift needs verified labels: the model learns each world's real rule, not grid
statistics.

These are mechanism numbers, not a leaderboard entry: a single 7B at 4-bit with one seed over
\ArcNumTTT{} held-out worlds, far below ARC solvers that add deep augmentation, voting, and
program search. The claim is narrow, and we think it is the important one: world-time compute---%
spending compute to learn verified world models---lifts generalization on real,
human-authored tasks, right where the ``you generated it'' critique bites hardest, with the
corrupted-label ablation as the test of whether the labels' exactness is what earns that lift.

\subsection{Across real domains and modalities---and where it stops (E80)}\label{sec:world-time-domains}
\begin{figure}[t]\centering
\includegraphics[width=\linewidth]{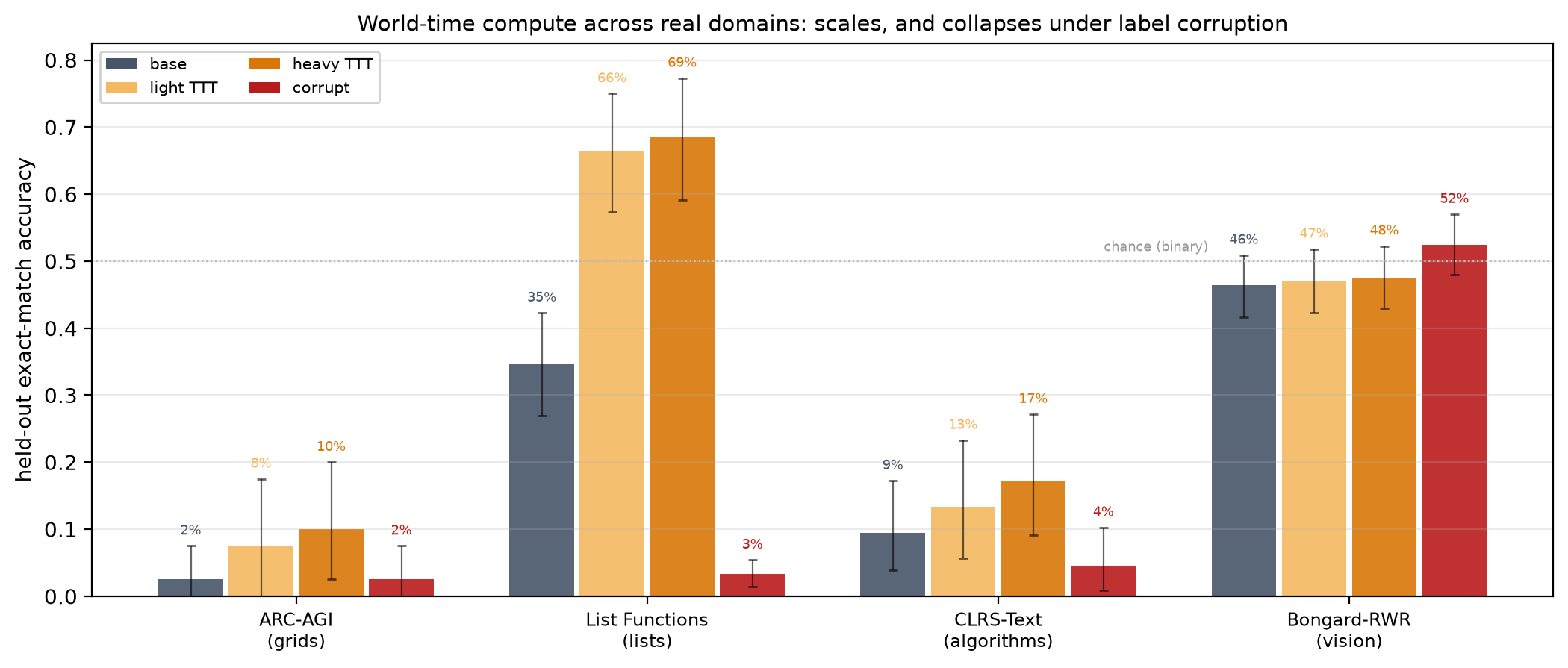}
\caption{\textbf{World-time compute across real domains and modalities (E80).} Per-world
test-time training (the same protocol everywhere) on real benchmarks: held-out exact-match for
base $\to$ light $\to$ heavy world-time compute, plus a corrupted-label control; error bars are
95\% bootstrap CIs over worlds. The three symbolic domains (ARC grids, List Functions lists,
CLRS algorithms) all scale with compute and collapse under corruption---the lift
needs verified labels. Bongard-RWR (vision, right) stays at chance: the boundary where the model must
induce a latent concept from rich perception, not from symbolic input. (The ARC bars
reproduce Fig.~\ref{fig:arc}, left, shown here for cross-domain comparison.)}
\label{fig:worldtime-domains}
\end{figure}

ARC is not a one-off. We ran the same per-world test-time-training protocol on two
more real benchmarks from different fields, plus one different \emph{modality}
(Figure~\ref{fig:worldtime-domains}): \textbf{List Functions}~\cite{srivastava2022bigbench}
(\LfNumWorlds{} hidden list$\to$list functions induced from input/output pairs),
\textbf{CLRS-Text}~\cite{markeeva2024clrstext,velickovic2022clrs} (\ClrsNumWorlds{} classic
algorithms---sorting, graphs, dynamic programming---each a hidden procedure to execute exactly),
and \textbf{Bongard-RWR}~\cite{bongardrwr2024,bongardrwrplus2025} (\BongNumWorlds{} real-image visual
concept-induction problems, via frozen DINOv2 features + a per-world head).

\paragraph{The trend replicates on every symbolic domain.} Held-out exact-match \emph{scales}
with world-time compute and \emph{collapses} to the floor once we randomize the demonstrations'
labels: List Functions \LfZero$\to$\LfLight$\to$\LfHeavy{} (heavy 95\% bootstrap CI over
worlds \LfHeavyCI), corrupt \LfCorrupt; CLRS \ClrsZero$\to$\ClrsLight$\to$\ClrsHeavy, corrupt
\ClrsCorrupt; ARC \ArcZeroShot$\to$\ArcTTTLight$\to$\ArcTTTHeavy{} (CI \ArcHeavyCI), corrupt
\ArcTTTCorrupt. Same mechanism, three independent real benchmarks, and in each one the labels'
exactness is what earns the lift---the discriminator from \S\ref{sec:discussion} holds everywhere.

\paragraph{Where it works: few-step reasoning.} The size of the effect tracks the
\emph{depth of reasoning} an instance needs. World-time compute pays off most where the
hidden rule is shallow---List Functions, a one-shot list transform, reaches \LfHeavy---and least
where it needs long multi-step execution---CLRS, whose answers are long algorithm traces, tops
out at \ClrsHeavy, the lowest of the three. This is the same boundary as the complexity cliff
(E20; mitigated by composition in the companion framework paper~\cite{openworld_framework}): verified and world-time-compute methods do best at tasks you can solve
in \emph{few reasoning steps} and fall off as the needed chain gets longer. It is also just
where they beat data-hungry learned world models---on the short-horizon \textbf{MiniGrid DoorKey}
planning task (\S\ref{sec:minigrid}) the verified model solves \emph{zero-shot} while DreamerV3
needs ${\sim}10^4$ interactions (\MGDreamerFirstSolve{} to first solve; E65) and V-JEPA-2 cannot control at all.

\paragraph{The boundary: rich perceptual abstraction.} Bongard-RWR is the honest negative: per-world
test-time training over frozen visual features stays at chance (base \BongBase, heavy \BongHeavy,
corrupt \BongCorrupt; Figure~\ref{fig:worldtime-domains}, right)---no scaling, no collapse. When
the model must \emph{induce} the latent concept \emph{from rich perception} instead of reading it from a symbolic
input, a small head over frozen features gets no grip. World-time compute is a lever for
symbolic, few-step reasoning, not a stand-in for perceptual representation learning---the same
scope boundary the paper draws throughout.

\subsection{Cross-world transfer on a real, human-authored domain (E84)}\label{sec:crossworld}
The real-data results above are per-world test-time training---the established
recipe~\cite{akyurek2024ttt}, not our contribution. The axis \openworld{} is built for is
cross-world generalization: train once on a family of verified worlds, then generalize to
held-out worlds. E83 found no such transfer across CLRS algorithms that share no skill.
E84 tests the flip side of that---that a \emph{coherent} family does transfer---on a real,
human-authored domain (List Functions). We train one LoRA adapter on \CrossWorldNTrain{} disjoint
held-in worlds, freeze it, and evaluate it zero-shot on \CrossWorldNEval{} held-out worlds it never saw
(\CrossWorldNSeeds{} seeds). Three arms share the same eval: \textbf{base} (no adapter),
\textbf{cross-world} (adapter trained with exact labels), and \textbf{corrupt} (the same held-in
prompts with randomized targets).

\begin{figure}[t]\centering
\includegraphics[width=0.6\linewidth]{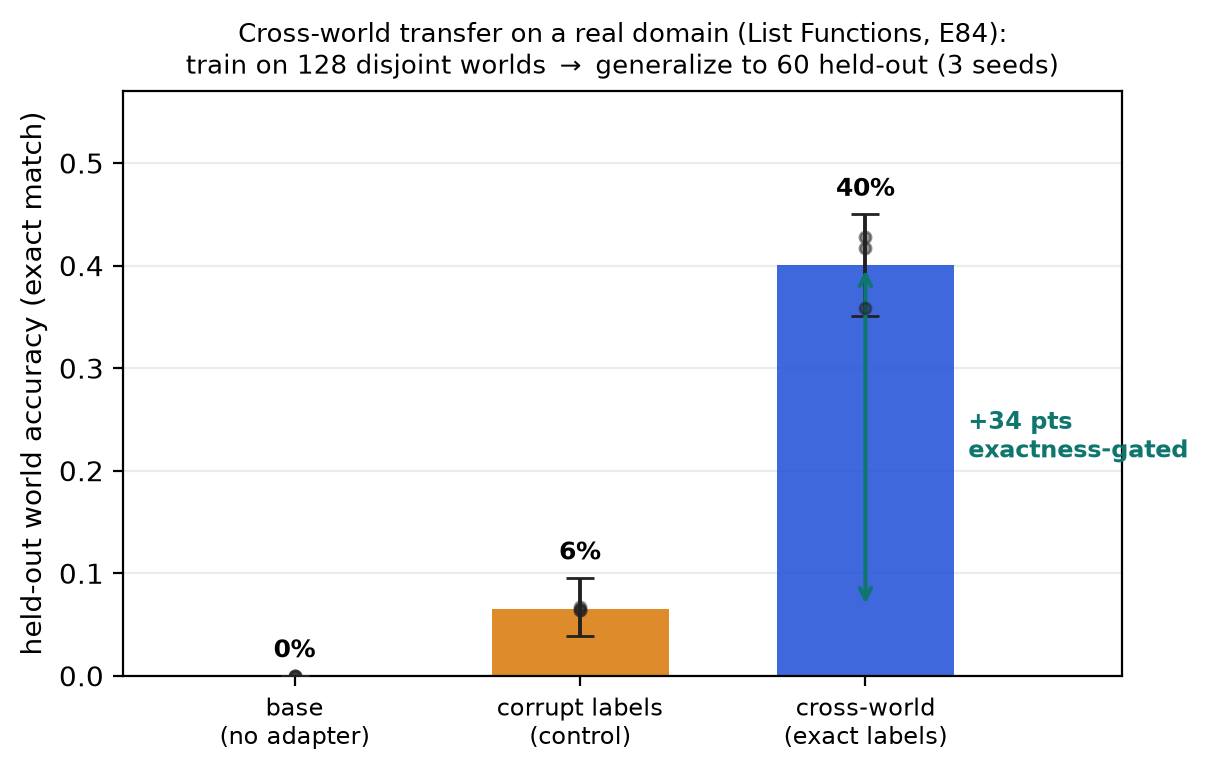}
\caption{\textbf{Cross-world transfer on a real domain (E84).} One adapter trained on
\CrossWorldNTrain{} disjoint List-Functions worlds generalizes to \CrossWorldNEval{}
held-out worlds it never saw (\CrossWorldMean, 95\% CI \CrossWorldCI), while the same training
with corrupted labels reaches only \CrossWorldCorrupt---a +\CrossWorldGap-point exactness-gated
lift (CI \CrossWorldGapCI). This is the cross-world axis, on data we did not generate,
that E83 (skill-disjoint algorithms) had left open.}\label{fig:crossworld}
\end{figure}

The cross-world adapter reaches \CrossWorldMean{} (95\% CI \CrossWorldCI) on worlds it never
trained on, versus \CrossWorldCorrupt{} (\CrossWorldCorruptCI) for the corrupted-label control---a
\textbf{+\CrossWorldGap-point} exactness-gated cross-world lift (CI \CrossWorldGapCI{}, which
excludes zero), and it holds steady across seeds (\CrossWorldSeeds{}; Figure~\ref{fig:crossworld}). The instruction-tuned
base scores \CrossWorldBase{} by exact match because it answers verbosely (chain-of-thought prose,
not a bare list). The corrupt control soaks up exactly that format effect: it learns the output
format from the same prompts, yet, with no correct rules, it gains only \CrossWorldCorrupt{}. So the gap
between cross-world and corrupt is \emph{rule-induction transferring across worlds}, not
formatting: an adapter that learned to infer list-function rules from one set of worlds infers
new rules in worlds it never saw. This is the novel axis---world-time compute as cross-world
generalization---shown on a real, human-authored domain, the result E83 left open.

\begin{figure}[t]\centering
\includegraphics[width=0.56\linewidth]{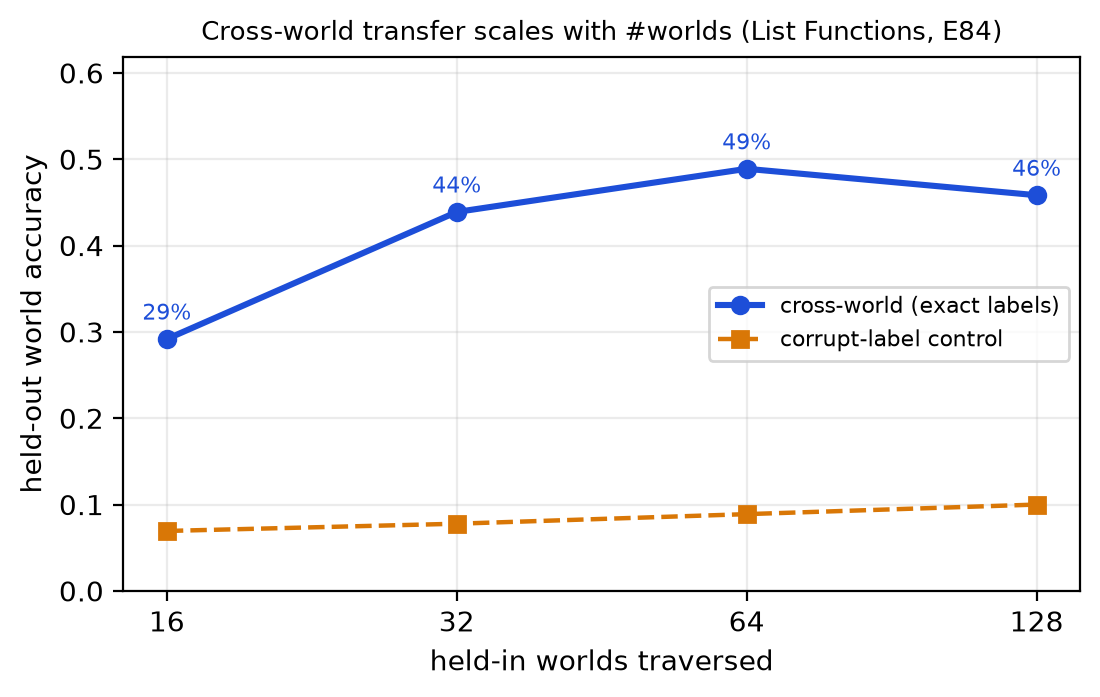}
\caption{\textbf{Cross-world transfer scales with \#worlds (E84, List Functions).} Held-out
accuracy rises with the number of disjoint held-in worlds traversed
(\CrossWorldLadderLo$\to$\CrossWorldLadderHi), peaking near 64 worlds then leveling off; the
corrupted-label control stays flat. (Single-seed ladder; the \CrossWorldNSeeds-seed headline of
Fig.~\ref{fig:crossworld} is \CrossWorldMean{} at \CrossWorldNTrain{} held-in worlds.) More verified
worlds $\to$ more transfer.}\label{fig:crossworld_ladder}
\end{figure}

\paragraph{It scales with worlds traversed.} The cross-world lift grows with the number of
held-in worlds---from \CrossWorldLadderLo{} at \CrossWorldLadderLoN{} worlds to a
\CrossWorldLadderHi{} peak by 64 worlds, then leveling off---while the corrupted-label control stays flat
(Figure~\ref{fig:crossworld_ladder}): a curve that rises then plateaus, for the novel axis, on real data.

\subsection{A descriptive regularity for world-time compute (E80)}\label{sec:world-time-law}
The effect is regular enough across E80 to state as a \emph{descriptive regularity}---not a
predictive law. We fit it over \LawNDomains{} domains with single-seed runs and one
out-of-sample check, the constants depend on the learner, and the fit is sensitive to which
domains we pick (a sibling fit over a different subset is anti-predictive). We report the shape
as a short summary, not a forecast. World-time compute follows a
\textbf{saturating learning curve}, $\mathrm{acc}(c)=\mathrm{base}+(C-\mathrm{base})(1-e^{-c/\kappa})$
(with compute-scale constant $\kappa$, distinct from the transition function $\tau$),
whose asymptote $C$ tracks \textbf{identifiability $\times$ realizability $\times$ headroom}
and which drops back to base without \textbf{exact labels}. We test it as a conjunction of
falsifiable clauses over \LawNWorlds{} worlds in \LawNDomains{} domains and three modalities
(symbolic text, vision, and tabular); what travels is the form, not point predictions.

\paragraph{Clause 1 (scaling): a small probe predicts the asymptote.} A light dose
of world-time compute predicts the heavy-dose asymptote: within-domain Spearman is \LawWithinArc{}
(ARC), \LawWithinListfn{} (List Functions), \LawWithinClrs{} (CLRS), \LawWithinBongard{}
(Bongard, vision), and \LawWithinTabular{} (tabular). A \emph{leave-one-domain-out} fit---train
the relation on the other domains, then predict the held-out one---reaches $R^2{=}\LawLodoRtwo{}$, and
the simplest model (light alone) beats adding parameters, which fits a real
regularity rather than an overfit. One out-of-sample check is consistent, though not decisive: the fit on grids, lists,
algorithms, and vision predicts per-world lift on a brand-new \textbf{tabular} domain it never
saw, at Spearman $\LawTabOOS{}$---one held-out modality, not a broad forecast.

\paragraph{Clause 2 (no free lunch): the predictor must be a probe.} We tried to predict the
asymptote from zero-training signals (teacher-forced answer log-probability against the number of
demonstrations). It fails---leave-one-domain-out $R^2{=}\LawFreeProxyRtwo{}$ (anti-predictive).
In-context likelihood taps Bayesian updating; the TTT lift taps weight-space learnability, and
the two pull apart. You cannot read the payoff off static signals; you have to spend a small dose of the
actual process. This negative result sharpens the regularity.

\paragraph{Clause 3 (learner-invariance): not just neural networks.} The same regularity holds for a
\emph{non-neural} learner---an enumerative program synthesizer over a list DSL, with search
budget as the compute axis: held-out exact-match rises and saturates (\SymScaleLo{} to
\SymScaleHi{} over the budget sweep), and corrupting the labels collapses it to \SymCorrupt{}
(no consistent program exists). Only which worlds are realizable depends on the learner (the
DSL is a different hypothesis class than the network); the curve and the exactness gate do not.
The regularity is about the verified-world learning problem, and it plays out the same way on gradient
descent and on symbolic search.

\paragraph{Clause 4 (falsification): it predicts its own nulls.} A regularity worth the name must say where it
fails. On the \ComboNWorlds{} deliberately heterogeneous worlds (E71) cross-world
transfer sits at the random floor; on ARC the cross-task ladder is flat; only coherent families
transfer---diagnosis (E74) and, on a real human-authored domain, List Functions
(E84, \S\ref{sec:crossworld}, a \mbox{+\CrossWorldGap}-point exactness-gated cross-world
lift)---and inside the heterogeneous set a coherence-guided search still
finds a generalizing subfamily (companion framework paper~\cite{openworld_framework}).
Here is a real-data probe of the same prediction (E83): train one adapter on 21 CLRS-Text algorithms
and evaluate 8 held-out algorithms, and you get no cross-world lift---held-out accuracy does
not rise above zero-shot---just as we expect for a set of algorithms (sorting, graph search,
string matching) that share little skill. We report this as a weak null, though: the
7B base sits near the floor on these held-out algorithms, so the test has little room to move in
either direction. One null we did not
predict in advance, and report as found rather than forecast: cross-language-frame transfer (E81,
\S\ref{sec:frame-invariance}) gives no lift, even though the exactness clause is consistent
with it (matched frames beat corrupted). The lesson is the honest one---the regularity's form
holds where we tested it, but a held-out language (or a CLRS algorithm) is its own skill,
outside the family structure the regularity scores.

\paragraph{The reader's guideline: how many worlds to simulate.} Because the curve saturates,
the best number of simulated worlds is its knee. From the per-world compute curve, $N_{90}$
(worlds for 90\% of the asymptotic gain) is \WCListfn{} for List Functions, \WCArc{} for ARC, and
\WCClrs{} for CLRS---\emph{harder domains need more simulated worlds}, but only a handful, not
thousands. Here is the scope: the curve's \emph{form} seems to recur across the verifiable-world problems
and learners we tested; the constants ($\kappa$, the depth cap) depend on the learner; and the predictor is
cheap but not free---a small probe, which puts the truly hard part where it belongs, in
choosing a representation that makes a world's rule identifiable.

\subsection{Composites with varying rules: language frames and the invariant principle (E81)}\label{sec:frame-invariance}
Every world so far has one rule-set. But you can write the same computational \emph{principle}
under many rule-sets, and world-time compute should recover the principle that stays the same across
all of them---a reference-frame view of generalization. Programming makes this concrete: a
\textbf{composite} world is one programming problem, and its \textbf{sub-worlds} are that same problem
written in \FrameNLangs{} languages (\FrameLangs)---each a verified code world whose transition is
that language's tests running. A language is a \emph{frame}: the same principle in different
coordinates. We build \FrameNWorlds{} such composites from HumanEval-X. We then ask a simple
question: does walking the other language-frames teach the frame-invariant principle well enough to
solve a held-out frame? For each problem and target language we measure three arms: zero-shot
(the base model on the held-out frame), \emph{frame TTT} (\FrameTttSteps{} steps of QLoRA on the
other languages' prompt$\to$solution pairs for that same problem, then evaluate the held-out
frame), and corrupt (the same budget spent on mismatched-language pairs---the exactness
control). We run two open-weight bases with different coding strength: Qwen2.5-Coder-7B (strong) and
Llama-3.1-8B (weak).

\begin{figure}[t]
\centering
\includegraphics[width=0.62\linewidth]{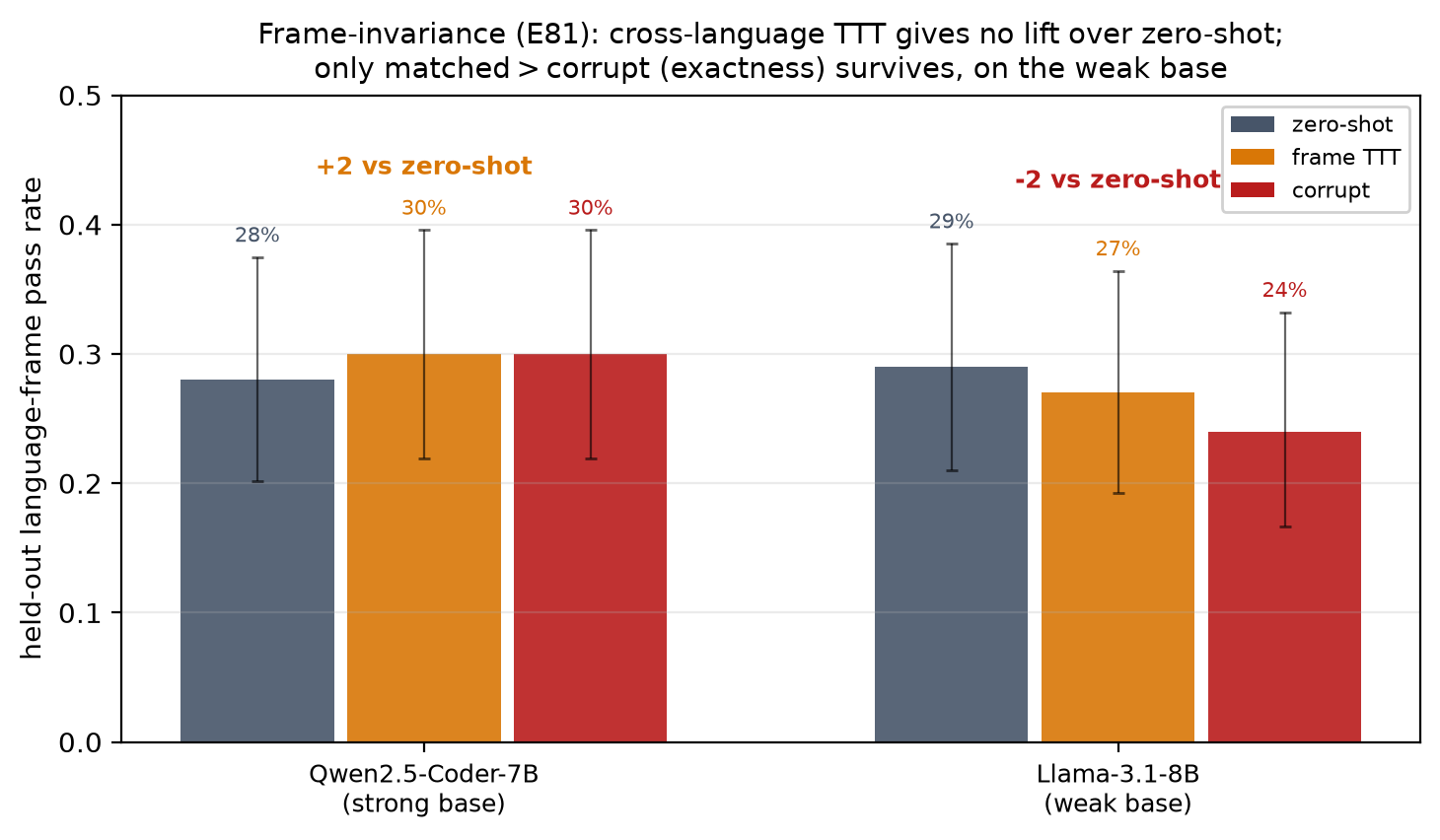}
\caption{\textbf{Frame-invariance across programming-language frames is a negative (E81).} A
composite world is one problem in \FrameNLangs{} languages; frame TTT fine-tunes on the other
languages and then evaluates a held-out language ($n{=}100$ problem$\times$language holdouts per base).
Cross-frame TTT does not beat zero-shot on either base: the strong coder stays flat
(\FrameQwenZero{}$\to$\FrameQwenTtt{}) and the weak base drops
(\FrameLlamaZero{}$\to$\FrameLlamaTtt{}); only the matched~$>$~corrupt ordering on the weak base
(\FrameLlamaTtt{} vs \FrameLlamaCorrupt) survives. We report this as a boundary of world-time
compute. Error bars are Wilson 95\% intervals.}
\label{fig:frame-invariance}
\end{figure}

This is the clearest negative in the paper, and we report it as a boundary
(Figure~\ref{fig:frame-invariance}). Cross-frame transfer does not beat zero-shot on either base.
The strong coder barely moves, \FrameQwenZero{}$\to$\FrameQwenTtt{} (within Wilson noise at $n{=}100$),
and the weak base---where the headroom hypothesis predicted the largest gain---instead
drops (\FrameLlamaZero{}$\to$\FrameLlamaTtt{}). The only structure that survives is a weak
exactness ordering on the weak base: matched frames beat corrupted ones (\FrameLlamaTtt{} vs
\FrameLlamaCorrupt). But even matched frames fail to reach the zero-shot baseline. So this differs from the
diagnosis-world (E74) and ARC (E80) settings, where world-time compute lifts held-out accuracy.
Walking the language-frames of one problem does not teach a transferable principle: a
held-out language acts like its own separate skill, and a few TTT steps on sibling languages disturb
the model more than they inform it. The honest lesson is a limit on the reference-frame analogy:
composites whose sub-worlds have genuinely different rule-sets (languages) are not automatically a
single learnable frame, and exactness (matched $>$ corrupt) is necessary but far from sufficient for
transfer. The clean positive on this substrate comes not from cross-frame TTT but from the
verified oracle used as a generator and backstop, which we isolate next (E82).

\subsection{The hybrid loop: a verified world as generator and backstop (E82)}\label{sec:hybrid-loop}
\begin{figure}[t]
\centering
\includegraphics[width=0.93\linewidth]{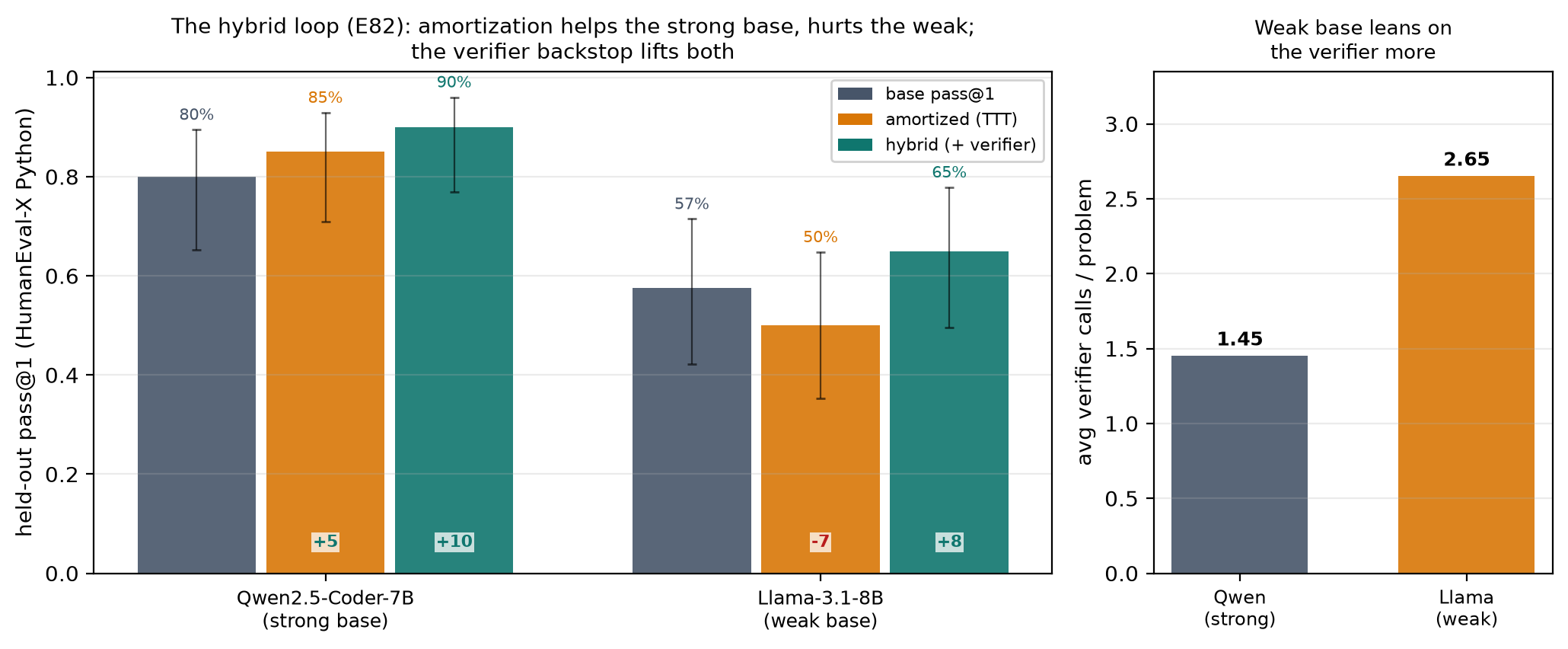}
\caption{\textbf{The hybrid loop with open weights and zero human labels (E82).} Left: held-out
pass@1 on HumanEval-X Python for base $\to$ amortized (QLoRA on self-generated test-passing
solutions) $\to$ hybrid (amortized model + verify-and-retry, up to \HybridRetry{}), with signed
deltas vs.\ base. Amortization helps the strong base (\HybridQwenAmortDelta) but hurts the weak one
(\HybridLlamaAmortDelta); the verifier backstop lifts both (\HybridQwenHybridDelta, \HybridLlamaHybridDelta).
Right: the weak base leans on the verifier more (\HybridLlamaCalls{} vs \HybridQwenCalls{} calls per
problem). Error bars are Wilson 95\% intervals over \HybridNEval{} problems.}
\label{fig:hybrid-loop}
\end{figure}

E81 shows that walking frames for free transfer fails on code; here the value of a verified world
is its exactness, used directly. E82 isolates that. It puts all three routes
(Table~\ref{tab:routes}) in one open-weight, zero-human-label experiment on HumanEval-X Python, whose
test suite is a perfect verifier. \emph{Route~1 (tool as generator):} sample from the base model and
keep only the solutions that pass the tests---exact training data with no labels (\HybridQwenPairs/\HybridNTrain{}
kept for Qwen, \HybridLlamaPairs/\HybridNTrain{} for Llama). \emph{Route~2 (amortize):} QLoRA-tune on
those verified solutions and measure pass@1 again. \emph{Route~3 (hybrid):} the amortized model
generates, the tool verifies, and on a failure it retries up to \HybridRetry{} times. We run the same
two bases as E81 over \HybridNEval{} held-out problems (Figure~\ref{fig:hybrid-loop}).

Three findings, one of them a caution. First, \textbf{amortization is not free}: self-distilling on
verified solutions \emph{helps} the strong base (\HybridQwenBase{}$\to$\HybridQwenAmort{},
\HybridQwenAmortDelta{} pts) but \emph{hurts} the weak one (\HybridLlamaBase{}$\to$\HybridLlamaAmort{},
\HybridLlamaAmortDelta{} pts)---the weak model's bootstrap set is thinner and noisier, so tuning on
it disturbs the model more than it teaches it. So the naive ``smaller models gain more'' prediction is
\emph{falsified} for pure test-time training on this task. Second, \textbf{the hybrid loop lifts
both} above base (\HybridQwenHybridDelta{} pts for Qwen to \HybridQwenHybrid{};
\HybridLlamaHybridDelta{} pts for Llama to \HybridLlamaHybrid{}): the verifier backstop recovers what
amortization alone puts at risk, and it is the robustly-positive route whatever the base strength. Third, the
headroom signal that does survive is \textbf{verifier reliance}: the weak base calls the tool
far more often inside the loop (\HybridLlamaCalls{} vs \HybridQwenCalls{} calls/problem). The
practical reading maps onto the three routes: amortization (Route~2) pays off only when the base is
strong enough to generate clean self-training data; the hybrid (Route~3) is the safe default; and a
weaker model simply pushes more of the work onto the exact world. The exactness that E45 et al.\ buy,
and that E81 found necessary-but-insufficient for transfer, is here \emph{sufficient}---when the world
is used as an oracle rather than a curriculum.

\subsection{Against trained dynamics}\label{sec:learned}
\begin{figure}[htb]
\centering
\includegraphics[width=0.97\linewidth]{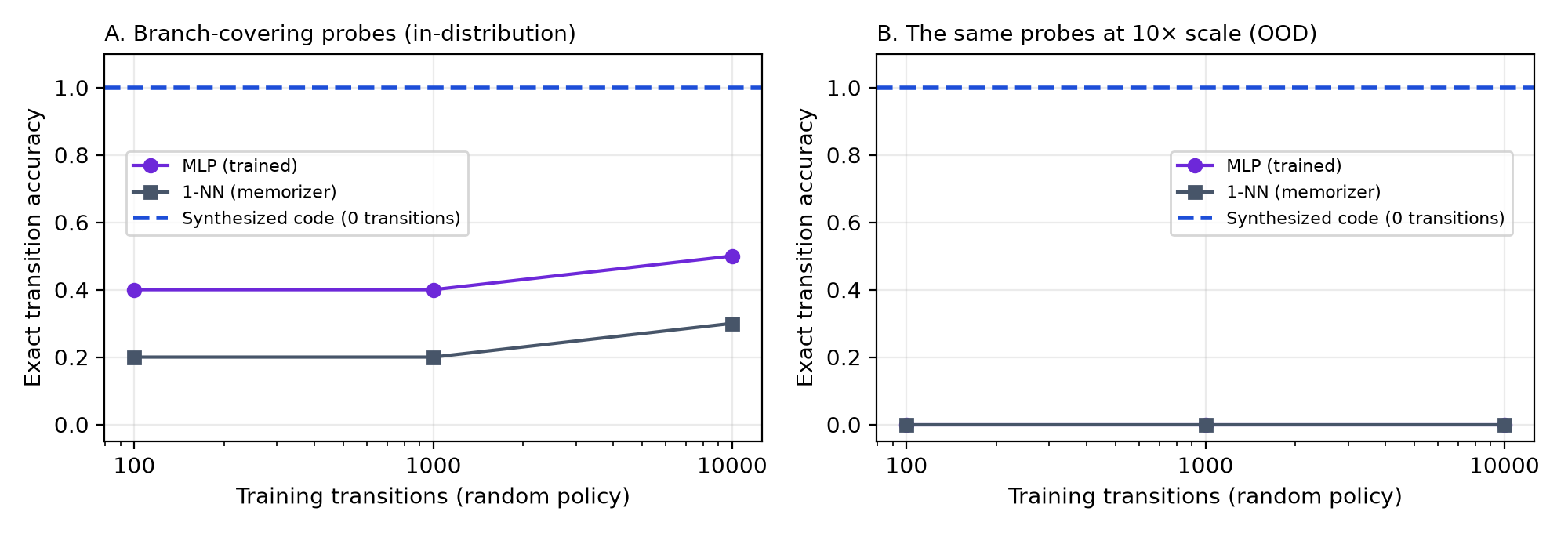}
\caption{\textbf{Trained baselines vs.\ the zero-transition program (E12).}
Exact transition accuracy of an MLP and a 1-NN memorizer trained on $K$
random-policy transitions of the sprint world, on branch-covering probes
in-distribution (\emph{A}) and at $10\times$ scale (\emph{B}). The dashed
line is the synthesized program, which uses no transitions at all. No trained
baseline transfers out of distribution.}
\label{fig:learned}
\end{figure}

The LLM next-state engine is a structural proxy; E12 adds real learned
baselines trained on environment transitions (Figure~\ref{fig:learned}). We
trained an MLP and a 1-NN memorizer, each on $K \in \{100, 1000, 10000\}$
random-policy transitions, and held them to the same evaluation as the synthesized
program. Two results stand out. First, sample efficiency: even at
$K=10{,}000$ the MLP reaches only \MLPInDistTenK{} exact accuracy on the
branch-covering probe suite, and \MLPOODTenK{} at $10\times$ scale; the
synthesized program scores 100\% on both from zero transitions---its
input is the rule text. Second, a measurement caution: at $K=10{,}000$ the
1-NN memorizer completes \KNNRolloutsTenK{} on-policy rollouts exactly while
scoring only \KNNInDistTenK{} on the probe suite. So on-policy rollout fidelity
can be \emph{memorized} without learning the rules. That is exactly why
branch-covering probes and OOD evaluation, not rollout agreement alone, are
the right instruments for dynamics quality.

E19 extends both axes (Table~\ref{tab:ladder-app}). A stronger
learned baseline---delta-state target, double capacity, lr-decayed
training---does no better than the original MLP (\DeltaMLPInDist{} in
distribution) and collapses in the same way out of distribution
(\DeltaMLPHundredX{} at both $10\times$ and $100\times$): the failure is not
under-tuning but extrapolation itself. Across a 1$\times$/10$\times$/100$\times$
scale ladder the synthesized program is exact everywhere
(\CodeHundredX{} at $100\times$); the LLM next-state engine is roughly
scale-insensitive but unreliable across the board ($\sim$40--50\% at every
scale)---it does not degrade OOD because it was never anchored to the
distribution in the first place.

\subsubsection{Induction on equal footing (E37)}\label{sec:induction}
A fair objection to the comparison so far is an \emph{information asymmetry}:
the synthesizer is handed the rules in natural language, while the learned
baselines must infer the dynamics from sampled transitions. E37 removes it
(Table~\ref{tab:induction}). The LLM sees only observed
$(\textrm{state},\textrm{action},\textrm{next})$ transitions---no rule
text---and must induce a \texttt{transition()} program, accepted only if it
reproduces the observed transitions; we then hold it to the same in-dist and
$10\times$ OOD probe suites as the MLP and 1-NN trained on the identical
data. Two findings result. First, on equal information, code
induction beats statistical learning out of distribution by a wide margin:
\IndTracesOod{} versus \IndMlpOod{} (MLP) and \IndKnnOod{} (1-NN). The
advantage is \emph{structural}: induced-code accuracy is the same
in-distribution and at $10\times$ scale (\IndTracesIn{} vs \IndTracesOod, on
every replicate), because the model induces symbolic rules that extrapolate,
whereas the learned baselines fit magnitudes and collapse to zero OOD on
every replicate. So the compounding/extrapolation advantage is not an
artifact of being handed the rules. Second, induction is \emph{imperfect}:
the 7B model recovers only \IndTracesIn{} of the probe behavior from traces
(it misreads the subtle \texttt{debt}-dependent interaction), well below the
rule-text anchor's \IndRulesIn. The gap between \IndRulesIn{} (rules) and
\IndTracesIn{} (traces) is what the specification
contributes---neither nothing nor everything---and it measures
the asymmetry the headline comparison carried. Larger budgets do not close
it here: the reachable transition set under a random policy saturates, so
$K{=}100$ and $K{=}1000$ give the inducer essentially the same distinct
examples.

\begin{table}[htb]\centering\small
\begin{tabular}{llcc}
\toprule
Method & Information given & In-dist. & $10\times$ OOD \\
\midrule
Code synthesis (rules) & rule text, 0 transitions & 1.00 & 1.00 \\
\midrule
\multicolumn{4}{l}{\emph{Equal information: 1000 random-policy transitions, no rule text}} \\
\textbf{Code induction} & traces only & 0.43 & 0.43 \\
MLP & traces only & 0.40 & 0.00 \\
1-NN memorizer & traces only & 0.17 & 0.00 \\
\bottomrule
\end{tabular}

\caption{E37, equal-information induction: exact probe accuracy when the LLM
must induce dynamics from transitions alone (no rule text), against the MLP
and 1-NN on the same data, plus the rule-text synthesis anchor. Code
induction is imperfect but scale-invariant (in-dist $=$ OOD); the learned
baselines collapse out of distribution; the rules-vs-traces gap measures what
the specification buys.}
\label{tab:induction}
\end{table}

\begin{figure}[htb]
\centering
\includegraphics[width=0.62\linewidth]{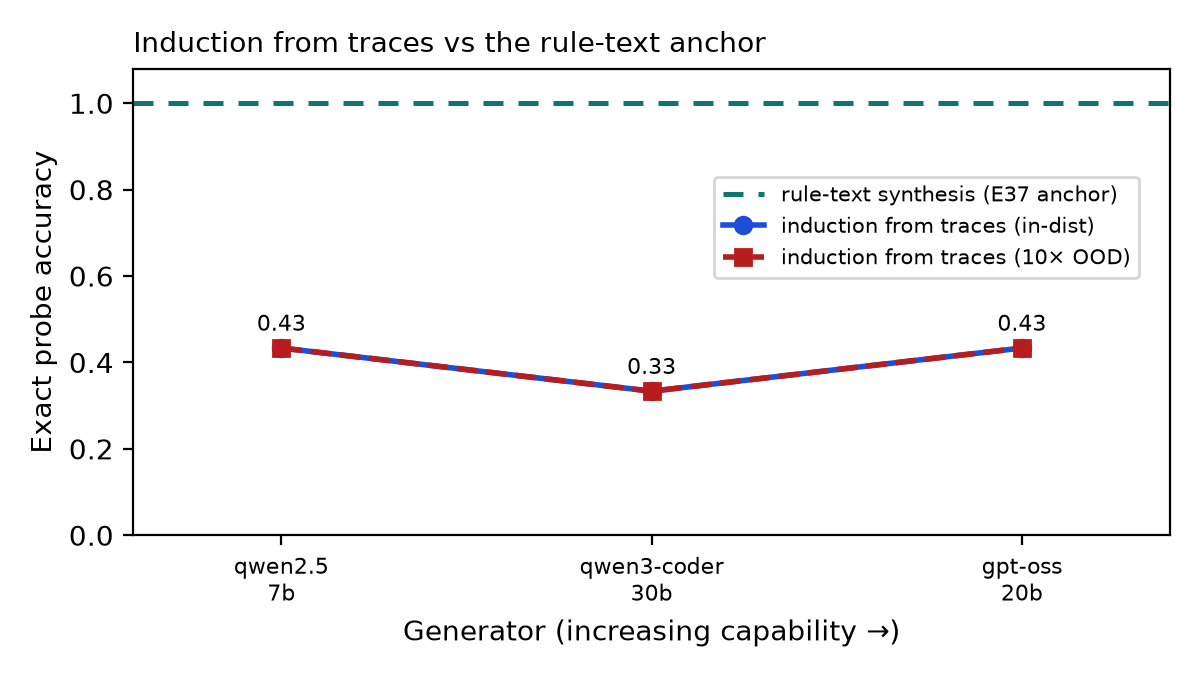}
\caption{\textbf{Induction does not scale with generator capability (E38).}
Trace-induction probe accuracy across three generator families; the rule-text
anchor (E37) is 1.00. The gap is roughly flat in model size and exactly
scale-invariant (in-dist $=$ OOD)---an identifiability limit of the data, not a
capability limit.}
\label{fig:induction-scale}
\end{figure}

\paragraph{Is the induction ceiling a capability limit? (E38)} If $\sim$0.43
reflects a weak 7B generator, a stronger model should close the rules-vs-traces
gap. It does not (Figure~\ref{fig:induction-scale}). Re-running E37's
trace-induction across \ScaleNumModels{} generators from three families
holds the gap roughly fixed: \ScaleQwenSmall{} for qwen2.5:7b,
\ScaleQwenCoder{} for the larger qwen3-coder:30b, and \ScaleGptoss{} for the
reasoning model gpt-oss:20b---none coming near the rule-text anchor's 1.00, and
every model exactly scale-invariant (in-distribution accuracy equals
$10\times$-OOD accuracy). Notably the 30B coder reproduces the observed
traces best ($\sim$0.9 of training transitions) yet generalizes no better;
it fits the shown examples without recovering the latent rule. So the ceiling is
an \emph{identifiability} limit of the data, not a capability limit:
random-policy transitions do not pin down the subtle interaction, so no
generator can recover it from them. Closing the gap needs informative
transitions (active experiment design), not a bigger model. (deepseek-r1:14b is
excluded: its reasoning-trace length made trace-induction impractical on the
local hardware.)

\subsubsection{Acting closes the gap that scale cannot (E43)}\label{sec:active-induction}
E38 names the bottleneck (identifiability, not capability) and points to the
fix: \emph{informative} transitions rather than a bigger generator. E43 tests
that fix directly. We cast dynamics recovery as version-space elimination over a
candidate family of \ActiveCands{} sprint rules that set the unknowns
(ship's debt increment, the divisor $k$ in the subtle $\textrm{bugs} \mathrel{+}=
\lfloor\textrm{debt}/k\rfloor$ interaction, and the fix/refactor magnitudes); a candidate
is dropped the moment its predicted next state disagrees with an observed
transition. We compare three probing policies on \ActiveRules{} hidden true
rules: a \emph{passive} random policy (the E38 setting); an \emph{active} policy
that picks the action that most splits the surviving candidates and tie-breaks
toward \texttt{ship} to build the debt that makes $k$ observable; and a
\emph{clairvoyant} reference that already knows the rule and greedily removes the
most candidates per step.

\begin{figure}[htb]
\centering
\includegraphics[width=0.92\linewidth]{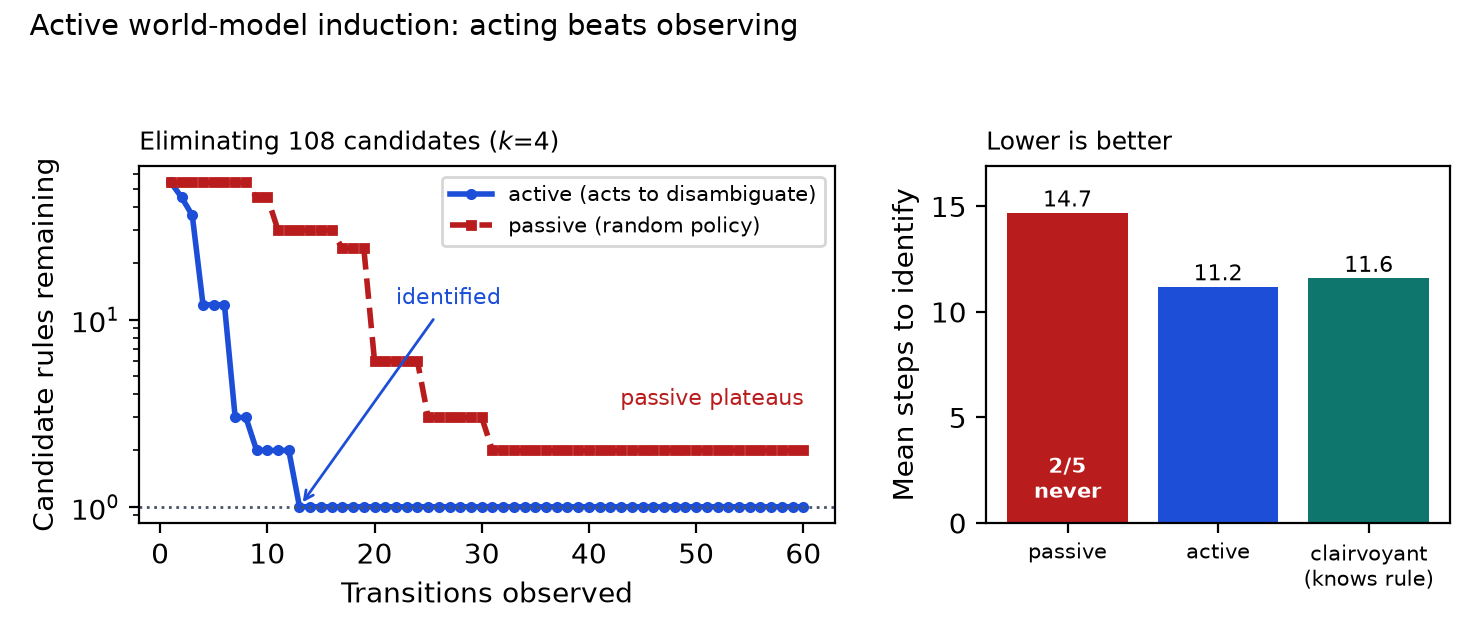}
\caption{\textbf{Acting beats observing for rule identification (E43).} Left:
candidate rules remaining (log scale) as transitions accrue, for a hidden rule
the passive policy never resolves---active elimination drives \ActiveCands{}
candidates to one while passive observation plateaus. Right: mean steps to
identify across \ActiveRules{} hidden rules; the active policy (knowing nothing)
matches the rule-knowing clairvoyant reference, while the passive policy is both
slower and leaves \ActivePassiveUnresolved{}/\ActiveRules{} rules unresolved.}
\label{fig:active-induction}
\end{figure}

Acting wins decisively (Figure~\ref{fig:active-induction}). The active policy
identifies the exact rule in \ActiveMeanSteps{} transitions on average, versus
\ActivePassiveSteps{} for the passive policy. And crucially, passive
observation never resolves \ActivePassiveUnresolved{}/\ActiveRules{} of
the hidden rules within the budget---exactly the high-$k$ rules whose
interaction only shows up at debt the random policy rarely reaches. This is the
E38 ceiling seen again as a plateau (Figure~\ref{fig:active-induction}, left):
no amount of passive data crosses it. The active policy, knowing nothing about
the rule, matches the clairvoyant reference within a step on average
(\ActiveMeanSteps{} vs.\ \ActiveClairSteps{}) and sometimes beats it---the
non-myopic ``build debt now to disambiguate later'' move is exactly what the
greedy reference, which optimizes per-step, fails to make. The lesson sharpens E38's:
when traces underdetermine the dynamics, the leverage is in \emph{choosing} the
transitions, and a world model the agent can act inside is what makes that
choice expressible.

\paragraph{Further world models on the same primitives.} Beyond induction, the framework can express many more world models that show off its range rather than carry the central argument: a non-enumerable version-space database (E46), path integrals over learning trajectories (E49), next-token world models with exact length generalization (E45), composite corporate (E48) and startup-growth (E51) worlds, a tree-of-thoughts brain simulator with real persistent memory (E58, E59), and the full perceive$\to$world$\to$emit$\to$act boundary (E60). These are the subject of the companion framework paper~\cite{openworld_framework}.

\subsection{From fidelity to decisions: planning}\label{sec:planning}
World models exist to improve decisions; E22 closes that chain
(Table~\ref{tab:planning}). A model-predictive planner runs exhaustive
depth-3 lookahead inside its world model and then executes the best action in the
ground-truth environment. Planning through the verified synthesized program
scores \PlanCode---the same as planning through the ground truth
itself (\PlanOracle), as bit-exactness guarantees---against \PlanReactive{}
for a reactive heuristic and \PlanRandom{} for random actions, with the
whole 12-step episode planned in \PlanCodeSeconds\,s. Planning through the
LLM next-state engine is held to depth 2 by cost (\PlanLLMSeconds\,s per
episode) and scores \PlanLLM---\emph{worse than no model at all}: lookahead
through hallucinated transitions steered the agent away from productive
actions entirely. We note one caveat: the LLM engine falls back to a
no-op when a reply does not parse as a next state, so this score mixes together
genuinely hallucinated transitions and unparseable outputs; the script now
logs the parse-failure rate so the two can be separated on a re-run, but the
qualitative conclusion---planning through an unverified model is worse than not
planning---does not depend on that split. Fidelity and throughput are not
abstract virtues; they are the difference between planning that helps and
planning that harms.

\begin{table}[htb]\centering\small
\begin{tabular}{lcc}
\toprule
Policy & Episode score & Planning time (s) \\
\midrule
\textbf{Lookahead d=3 via synthesized code} & 7.5 & 0.03 \\
Lookahead d=3 via ground truth (bound) & 7.5 & 0.00 \\
Lookahead d=2 via LLM next-state & 0.0 & 85.85 \\
Reactive heuristic (no model) & 5.0 & 0.00 \\
Random policy (5 seeds, mean) & 3.8 & --- \\
\bottomrule
\end{tabular}

\caption{E22: model-predictive lookahead on the sprint world, plans executed
in the ground-truth environment. Value function: shipped $-$ bugs $-$
0.5$\cdot$debt after 12 steps.}
\label{tab:planning}
\end{table}

\subsection{Crossing species: verified vs.\ trained vs.\ perceptual world models (E63, E65, E67)}\label{sec:bakeoff}
The comparisons so far pit verified code against an LLM next-state proxy; a model-based-RL reader
asks how it fares against world models trained on transitions, judged on task return. Across
\BakeMethods{} locally-runnable world models on \BakeDomains{} symbolic domains under one battery
($K{=}10{,}000$ transitions, 5 seeds; Table~\ref{tab:bakeoff}), the result is uniform: the verified
code model is exact in- and out-of-distribution, rolls out bit-exactly from zero transitions,
and plans optimally (return~\BakeCodeReturn) at \BakeCodeSpeed{} steps/s, while every learned model
drops to \BakeLearnedOodMax{} exact accuracy at $10\times$ OOD and plans at or below model-free
control---sometimes negative (planning through a trained model is worse than not planning, as
lookahead drives the agent into unseen states; E61). Perceptual world
models---DreamerV3~\cite{dreamerv3}, MuZero~\cite{muzero}, IRIS~\cite{iris2023},
DIAMOND~\cite{diamond2024}, Genie~\cite{genie2024}, Sora~\cite{sora2024}, V-JEPA~\cite{vjepa2025}---are
a different species (they predict observations or embeddings, not symbolic state), so we
compare them on \emph{properties} (Table~\ref{tab:bakeoff}, below), not a fabricated shared number.

\begin{table}[htb]\centering\small
\begin{tabular}{llccccc}
\toprule
World model & Train data & Probe & Probe & Rollout & Control & Audit. \\
& (transitions) & in-dist & 10$\times$OOD & exact & return & \\
\midrule
verified code (CWM) & 0 (rules) & 1.00 & 1.00 & 1.00 & +7.5 & \checkmark \\
1-NN & 10,000 & 1.00 & 0.00 & 0.77 & +5.7 & -- \\
tabular & 10,000 & 1.00 & 0.00 & 0.77 & +5.7 & -- \\
linear & 10,000 & 0.46 & 0.00 & 0.19 & -7.4 & -- \\
koopman & 10,000 & 0.75 & 0.00 & 0.44 & +5.6 & -- \\
MLP & 10,000 & 0.70 & 0.00 & 0.19 & -2.6 & -- \\
LLM next-state$^\dagger$ & rules & -- & -- & 0.00 & +0.0 & -- \\
\bottomrule
\end{tabular}

\caption{\textbf{World-model bake-off (E63).} Every world model we can run locally on shared
symbolic tasks (5 seeds, $K{=}10{,}000$). Verified code is the only model exact, OOD-robust,
optimal-for-control, auditable, and zero-data; perceptual models are a different species
(perceptual, not symbolic), positioned against \openworld{} in the introduction.}
\label{tab:bakeoff}
\end{table}

\paragraph{Where the species can meet (E65 MiniGrid, E67 cartpole).}\label{sec:minigrid}\label{sec:cartpole}
On benchmarks where learned models are strongest---image-based MiniGrid DoorKey and pixel
cartpole-swingup, each with OpenWorld's transition validated bit-for-bit against the real
environment---the picture holds on the learned models' own terrain. OpenWorld solves DoorKey
zero-shot (a \MGOpenWorldPlan-step optimal plan, no data), where DreamerV3 needs $\sim$10k pixel
interactions just to succeed once; on cartpole it solves \CartOpenWorldSolve{} of episodes with
zero data via CEM-MPC, while a frozen-perceptual V-JEPA-2 fails through every controller we tried (a
control failure, not a perception failure). These are scope properties, not contests---handed the
dynamics, the verified model need not learn them; the full three-species tables are in the companion
framework paper~\cite{openworld_framework}. The boundary is consistent: verified world models win
where the \emph{reasoning}, not the perception, is hard and the horizon is short (DoorKey), and stall
where rich perceptual abstraction is the task (Bongard-RWR, \S\ref{sec:world-time-domains}).

\subsection{Scope boundaries and oracle-free verification}\label{sec:boundaries}
Three round-3 experiments chart where the paradigm holds and what to do at
its edge.

\begin{figure}[htb]
\centering
\includegraphics[width=0.6\linewidth]{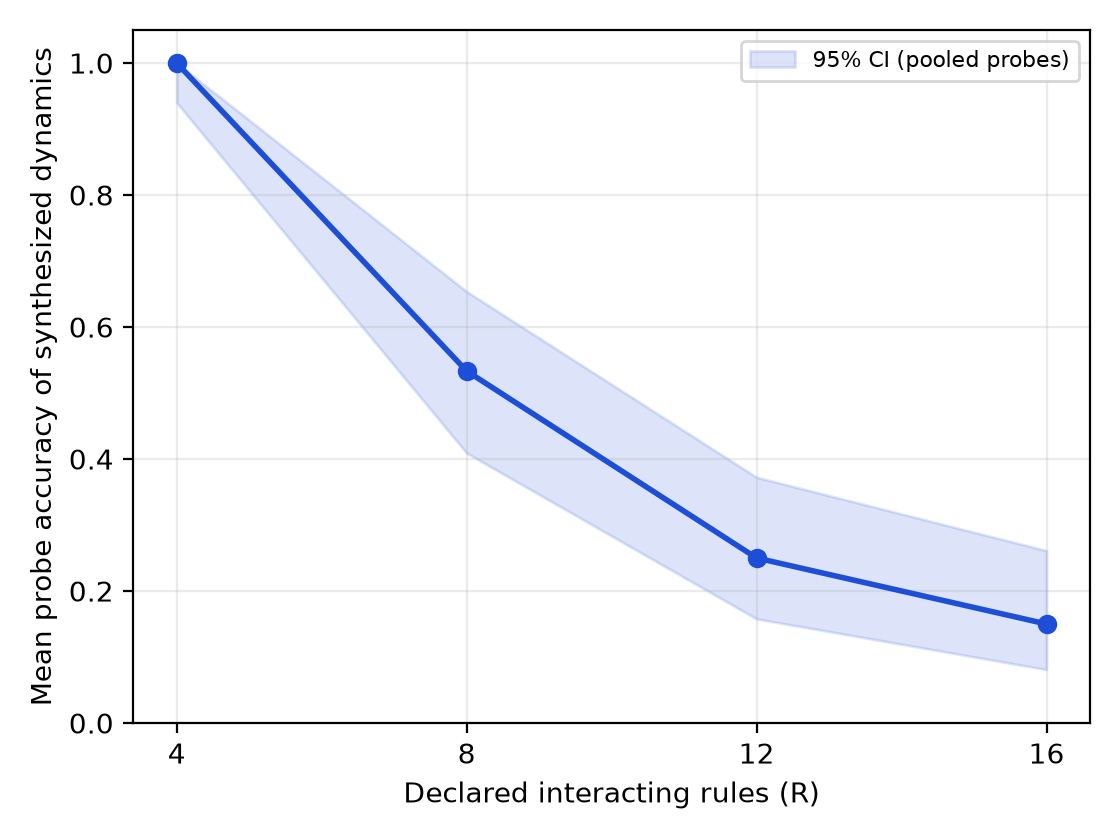}
\caption{\textbf{The complexity cliff (E20).} Probe accuracy of 7B-synthesized
dynamics versus the number of declared interacting rules; three syntheses per
point, shaded band is the 95\% CI over pooled probes. Monolithic synthesis is
reliable at $R{=}4$ and unreliable past $R{\approx}8$.}
\label{fig:complexity}
\end{figure}

\paragraph{The complexity cliff (E20).} Parametric worlds with $R$
interacting rules (conditions reading pre-action state, summed effects,
clamping) were synthesized from rule text at $R \in \{4, 8, 12, 16\}$
(Figure~\ref{fig:complexity}). Mean probe accuracy falls from \CplxFour{} at
$R{=}4$ to \CplxEight{} at $R{=}8$, \CplxTwelve{} at $R{=}12$, and
\CplxSixteen{} at $R{=}16$: monolithic single-function synthesis with a 7B
generator degrades sharply once roughly eight coupled rules must be
written at once. This is the paradigm's present boundary at this model
scale, and it motivates compositional synthesis---one expert per rule, as in
PoE-World~\cite{poeworld2025}---as the structural fix.

\paragraph{Stochastic dynamics (E21).} Threading a seed through the state
(\texttt{rng\_seed} in, derived seed out) extends verified synthesis to
stochastic worlds without giving up replayability. On a queueing world
with a declared 0.3 arrival probability, accepted programs (\StochAccept{}
of attempts) matched the declared distribution to within
\StochArrivalErr{}~percentage points over 2{,}000 seeded transitions, kept
the deterministic bookkeeping perfect (\StochDetAcc), replayed bit-exactly,
and in fact matched the hand-written stochastic oracle bit-for-bit
(\StochOracleExact).

\paragraph{Oracle-free error detection (E23).} Practitioners lack the
hand-written oracles our evaluations use; they can, however, synthesize the
dynamics several times. Across \SelfCheckPrograms{} stored programs and
\SelfCheckPairs{} program--probe pairs, disagreement with the ensemble
majority caught ground-truth errors with \SelfCheckPrecision{} precision
and \SelfCheckRecall{} recall---the majority vote rebuilt the oracle
exactly on these worlds---and a program's agreement rate ranked its true
accuracy perfectly (Spearman $\rho=\SelfCheckSpearman$). The caveat is
structural: majority-as-oracle fails under correlated errors (e.g., every
model misreading the same ambiguous rule), so ensemble disagreement is a
cheap screen, not a proof; it composes naturally with the invariant and
critic gates.

\section{Discussion and Limitations}\label{sec:discussion}
\paragraph{Interpretation.} The experiments measure a structural trade-off.
Learned world models spend training compute to cover pixel-native breadth, and they pay
for it with compounding error, opacity, and frozen values. Verified code synthesis
buys exactness, speed, OOD transfer, and auditability, and it pays at the
boundary of what you can state symbolically. Inside that boundary the
results are one-sided: a laptop-scale local model, run as
plan--generate--verify, builds simulators that are \emph{perfect} where
learned-style baselines are only approximate. The ablations license a plain workflow:
state the rules precisely; verify them with invariants; test the accepted
artifact against what you meant; then tighten and recompile. Each gate buys measurable
correctness, and the artifact stays a diffable program at every stage.

\paragraph{Is world-time compute just fine-tuning?} Mechanically, yes: there is no new
optimizer or architecture, only LoRA-SFT. The claim is about the \emph{data source}, not
the optimizer. Ordinary fine-tuning needs a labeled corpus---scarce, costly, and sometimes
wrong; world-time compute instead fine-tunes on trajectories from a \emph{verified} world
model, which makes unlimited examples and labels them \emph{exactly} for free, because the
transition is code and the oracle is running it. In one line: a correct world model turns
compute into unlimited correct training data. Is the ``verified'' qualifier really
load-bearing---or just decoration on plain SFT? That is an empirical question with a clean
discriminator, the \textbf{corrupted-label ablation}: randomizing the labels keeps the
fine-tuning and the world's structure but destroys their correctness, so if accuracy does not
change, the framing adds nothing. It does collapse to the no-fine-tune floor: the
corrupted-demonstration arm on real ARC worlds falls to \ArcTTTCorrupt{} against the verified
arm (\S\ref{sec:world-time-arc}, Fig.~\ref{fig:arc}). So the contribution
is conceptual, not algorithmic, and only two
things earn the approach its name over ``SFT with extra steps'': (i) this verified-label
dependence, and (ii) cross-world transfer---training on a family of worlds lifts
accuracy on held-out worlds never fine-tuned on (E74, E76), which same-world
fine-tuning cannot do. Where neither holds---heterogeneous worlds with no shared skill
(E71, E73), or a benchmark of per-task-novel rules where cross-task transfer stays weak
(ARC, \S\ref{sec:world-time-arc})---it is, fairly, just expensive fine-tuning, and we say so.
This puts world-time compute squarely in the self-training lineage---%
STaR~\cite{star2022}, ReST and ReST-EM~\cite{rest2023,restem2024}, expert
iteration~\cite{expertiteration2017}, rejection-sampling fine-tuning~\cite{rft2023}, and
learning from verifiable rewards~\cite{rlvr2024}---all of which retrain a model on its own
outputs filtered by a correctness check. Two differences are the contribution, not the loop:
the verifier is a \emph{full world model with dynamics}, so the training unit is a verified
trajectory (a state-action-next-state transition), not a checked final answer; and the
target is a different axis of generalization---held-out worlds in a family, rather
than the same task improved in place. In those terms the corrupted-label ablation is exactly
a verifier-reliability study: the failure mode (training on confidently-wrong self-labels) that
self-training fears most, here ruled out by construction.

\paragraph{The trajectory unit, measured (E85).} We test the first difference---the unit is a
trajectory $(s,a,s')$, not an answer---directly, on synthetic iterated worlds: one
cross-world adapter trained on verified single-step transitions vs.\ another trained on
$(s_0\!\to\!\text{final})$ answer pairs from the same worlds (\TrajNSeeds{} seeds). At the
trained horizon the trajectory unit wins (\TrajTrainTraj{} vs.\ \TrajTrainAnswer{} for the answer
unit, $+\TrajTrainGap$ points)---the unit, not the loop, carries the signal. Rolled out to a
horizon \emph{longer} than trained, the answer unit catches up (\TrajLongAnswer{} vs.\
\TrajLongTraj)---the model's per-step predictions compound, the very error mode verified
code avoids. So the trajectory is the better training signal, but composing it at inference
brings compounding back on the learned side.

\paragraph{When does it pay off? Few-step reasoning.} Pulling the results together
(\S\ref{sec:world-time-domains}) gives a single operating regime. Verified world models and
world-time compute win where a task boils down to a few reasoning steps over a state that can
be \emph{told} symbolically: shallow rule induction (List Functions, the largest lift), small
grid transforms (ARC), and short-horizon planning (MiniGrid DoorKey, solved zero-shot while
DreamerV3 needs $10^4$--$10^5$ interactions~\cite{dreamerv3}). Performance drops along two axes---as the
required reasoning \emph{lengthens} (CLRS algorithm traces, the lowest of the symbolic domains;
and the E20 complexity cliff past $\sim$8 composed rules), and as the latent rule must be
induced from rich perception rather than read from symbolic state (Bongard-RWR vision,
flat at chance). This is not a hedge but the shape of the contribution: a verified, auditable,
zero-data lever that is strongest exactly where today's data-hungry learned world models are
weakest---short, specifiable reasoning---and that cleanly hands the long-horizon, perception-heavy
regime to them.

\paragraph{Three routes to use a world model.} A verified world model is a reusable artifact, and
it is useful whether or not you ever fine-tune; world-time compute is one of three ways to spend
it (Table~\ref{tab:routes}). \emph{Route 1---as a tool:} serve the world and call it---plan
through it, ask for exact next-states, or use it as a verifier (no training; exact, auditable;
\S\ref{sec:planning}, \S\ref{sec:minigrid}). \emph{Route 2---per-world test-time training:}
distill a single world's exactly-labeled trajectories into weights with QLoRA, amortizing that
world's skill for tool-free inference (\S\ref{sec:world-time-arc}). \emph{Route 3---world-time
compute:} traverse a \emph{family} of verified worlds and fine-tune to generalize to held-out
worlds (\S\ref{sec:world-time}, \S\ref{sec:world-time-law}). The \emph{hybrid} combines them: use
the tool to \emph{generate} exact trajectories, train to \emph{internalize} them, and keep the
exact tool as a fallback for high-stakes queries---the tool bootstraps the data, training
amortizes it, and exactness is recoverable on demand.

\begin{table}[t]\centering\small
\resizebox{\linewidth}{!}{%
\begin{tabular}{lcccc}
\toprule
Route & Exact? & Training & Tool at inference & Generalizes to new worlds \\
\midrule
1.\ Tool & yes & none & yes & n/a (use any world) \\
2.\ Per-world TTT & approx. & QLoRA, one world & no & within the world \\
3.\ World-time compute & approx. & QLoRA, many worlds & no & yes (across the family) \\
Hybrid & on fallback & QLoRA & only when unsure & yes, with exact backstop \\
\bottomrule
\end{tabular}
}
\caption{\textbf{Three routes to use a verified world model}, plus the hybrid. Need
exactness/audit $\to$ Route 1; amortize one world $\to$ Route 2; generalize a family $\to$
Route 3; best of both $\to$ Hybrid.}
\label{tab:routes}
\end{table}

\paragraph{Value alignment as an open specification.} The dial experiments
ground the alignment claim: because objectives are declared artifacts, the
welfare--fairness frontier is operational---an operator moves along it
in real time, and a tuner searches moral configurations jointly with world
design, surfacing the whole manifold of value settings that solve a task rather
than a single frozen compromise. Two qualifications, owed to the
philosophical review. First, the openness reaches only as far as the expressible
structures: a weighted sum is act consequentialism, and
the companion framework paper~\cite{openworld_framework} shows that aggregator choice, side constraints,
and theory uncertainty are distinct moral objects the framework must (and
now does) represent rather than approximate with weights. Second, open
specification \emph{manages} the specification trap~\cite{spectrap2025}
procedurally rather than resolving it: the trap moves from frozen
weights to the operator's hand on the dial, and which hand that should
be---justifiability to those affected, in the contractualist's
sense~\cite{scanlon1998}---is a question no architecture settles. The
tuner's solving manifold is at best a crude contractualist surface: the set
of configurations that meet every stakeholder's declared minimum.

\paragraph{Limitations.} (i) Scope: symbolic-state worlds; pixel-native
domains stay the territory of learned models---the comparison here is
within the symbolic regime, and neither the LLM proxy nor the numpy
baselines is a trained Dreamer. The headline comparison also hands the
synthesizer the rule text; E37 shows the structural OOD advantage survives
removing that asymmetry (induction from traces alone), but induction is then
imperfect (\IndTracesIn{} vs the rule-text \IndRulesIn), so part of the
headline margin reflects the value of an explicit specification, not the
representation alone. (ii) Complexity: monolithic synthesis with a
7B generator is reliable at four interacting rules and unreliable past
roughly eight (E20); the headline worlds sit below that cliff, and richer
worlds will need compositional synthesis~\cite{poeworld2025} or larger
generators. (iii) Scale: twenty repair tasks and three handwritten worlds;
paired tests sharpen the comparisons, but the benchmark is still an archetype
suite, not HumanEval~\cite{humaneval2021}, 3B+ agents saturate it, and the
classic defect archetypes are plausibly present in pretraining data: agent
solve rates should not be read as measurements of debugging skill---the
world-model claims are unaffected, since the environment runs exactly
regardless of what the agent has memorized. (iv) Synthesis replicates across
three model families (E16), but the judge, critic, and repair-agent roles
stay Qwen-only. The synthesis, judge, and repair experiments run on one
consumer machine under the stated quantizations; the fine-tuning experiments
(E74, E78, E80--E83) instead run on a single cloud A100 (4-bit QLoRA for the 7B
and larger models), which we state plainly so the consumer-hardware framing is
not read as covering the training runs. (v) Judges: selection accuracy and order-consistency are
measured but imperfect, the pooled selection margin is marginal
($p=\PooledMcNemarP$) and costs $\sim$4$\times$ inference, and rubric grading
showed compression; judge verdicts should gate, not replace, programmatic
verification. (vi) Ensemble self-checking (E23) reached perfect
precision/recall here but is structurally vulnerable to correlated errors
across generators reading the same ambiguous rule; it is a screen, not a
proof. (vii) The sandbox is best-effort isolation against accident, not an
adversarial security boundary (generated patches twice defeated in-process
timeouts before the fork-based runner; see the repository history).
(viii) The moral machinery, even broadened, evaluates declared values; whose
values are declared---and the legitimacy of the declaring hand---is a
procedural question outside any architecture~\cite{scanlon1998}, and the
parliament's credences are themselves operator-set dials. (ix) Several
experiments were redesigned after pilots and four internal review rounds
(E3, E5/E6, E11--E27); pilot saturation results are reported alongside the
final protocols, and all reviews ship in the repository.
(x) Multiple comparisons: across \NumExperiments{} experiments we report many
$p$-values and rank correlations and apply no family-wise or
false-discovery correction; with this many tests the borderline results
($p$ between $0.05$ and $0.10$---notably the judge margin and the
rubric-paraphrase correlation) should be read as exploratory, and only the
large deterministic effects (exact-match rates, OOD collapse, the complexity
cliff) hold up under the multiplicity. (xi) Variance: the deterministic claims
are exact-match against oracles and need no error bars, but several
\emph{stochastic} (LLM) headline quantities are point estimates from small,
single-seed samples (e.g.\ the E1 divergence step, $n{=}3$, one world) with no
across-seed variance; we add the pooled multi-world view (E11) where it
matters, but per-model probe accuracies should be read as estimates, not tight
measurements. This caveat extends to the headline world-time-compute results: the
E74 model-size scaling is a single fine-tuning seed per size, and the real-data
ARC/CLRS arms are single-seed at $n{\approx}30$--$40$ with CIs that touch the floor,
so the size-trend and the weaker real-data domains should be read as exploratory;
for ARC we report a \ArcSeedsN-seed re-run---heavy \ArcHeavySeedMean{} (seeds
\ArcHeavySeedList) vs.\ corrupt \ArcCorruptSeedMean---so the verified-vs-corrupt gap holds across
seeds, though absolute scores stay mechanism-scale. (xii) Two illustrative studies are true \emph{by construction}:
E59's architecture lift follows arithmetically from a modeled fixed-capability
backbone, and E60's routing gap over exact-key lookup is guaranteed by the
comparison design (we add a fair lexical baseline and report the modest gap over
it); both are flagged in place as engineering demonstrations, not measured effects.
(xiii) Porting to real data surfaces two validity conditions: the input representation must
encode per-world structure (a ProteinGym port describing each mutation only by global
biochemical deltas made every assay-world identical and flattened the world-count curve until
per-world local-sequence context was restored), and faithful data loading---e.g.\ the wild-type
sequence and one-based mutation indexing---is the dominant correctness risk. We treat both as the
first checks any real-domain port must pass.
(xiv) The trajectory-vs-answer distinction is isolated in E85 (above) on synthetic iterated
worlds; porting that head-to-head to a multi-step real domain (where trajectory and answer
diverge most) remains future work.

\section{Conclusion}
A training-free symbolic world model---synthesized as code by a local model and verified
before acceptance---is bit-exact where learned-style dynamics compound error, and plans
\emph{as well as the ground truth itself} where planning through a hallucinating model is
worse than no model. And beyond serving as a simulator to plan in, the same
verified worlds are training signal: traversing many world models of a domain
lifts a model's generalization to held-out worlds, scaling with the number of worlds
traversed---world-time compute, a training-time analogue of test-time compute.
Its boundary is measured just as sharply: monolithic 7B
synthesis fails past roughly eight interacting rules. The world-model field's
path has been to scale parameters until hallucination is suppressed;
these results argue for an orthogonal path---make the model a program, verify
the program, and keep the values dialable. Next steps: compositional
synthesis to cross the complexity cliff~\cite{poeworld2025}, the coding world
at standard-benchmark scale, hybrid perception-to-symbol pipelines, and
judges that propose---not just select---world repairs.

\section*{Declaration of generative AI use}
During the preparation of this work the authors used \emph{Claude 4.8}
(Anthropic, via Claude Code) to implement the \openworld{} framework and
experiment scripts, run the experimental campaign, generate figures and
tables, and draft and edit this manuscript and the experiments, under the
authors' direction. The authors reviewed and edited the content and
take full responsibility for the content of the publication. All experimental results are produced by the released
deterministic scripts (not by generative-AI estimation), and no AI tool is
listed as an author, since such tools do not meet authorship criteria.

\section*{Data and code availability}
All code, the benchmark suite, experiment scripts, raw result JSON, and this
manuscript's build are available at
\href{https://github.com/quome-cloud/openworld}{\nolinkurl{github.com/quome-cloud/openworld}}.
All data are synthetic; no human-subject or patient data were used.

\section*{Author contributions}
J.S. conceived the study, directed the implementation and experiments, and
reviewed the manuscript. I.S., J.R., M.O., C.O., C.K., A.E., M.B., R.T., J.T., and M.G.F.
contributed to the research and edited the manuscript. Many of the authors are
active contributors to the \openworld{} open-source project: they ran their own
simulations with the framework and provided feedback that improved the product.
All authors reviewed and approved the final manuscript.

\section*{Competing interests}
The authors declare no competing interests.

\section*{Funding}
This work received no specific external funding.

\appendix
\section{Index of experiments}\label{app:experiment-index}
The experiments E1--E85 are introduced across the Results section in thematic
rather than numeric order; this index lists each by number with a one-line
description and the section, figure, or table where it appears, as a navigation
aid. Numbers are non-contiguous: E14 and E53 are retired (absorbed or cut) and
do not appear in the text; E28--E29 are described in the Experimental Setup but
their dedicated results section was cut; and E66, E69--E73, E75, E78 (its E78b
variant \emph{is} used), and E79 were not used (a superseded probe or a negative
result excluded from the paper).
The world-time-compute family E74--E85 (E80 spans several real-data domains and
the descriptive regularity) carries the central result.

\begingroup\footnotesize\setlength{\tabcolsep}{4pt}
\begin{longtable}{@{}r p{0.62\textwidth} l@{}}
\caption{Index of experiments E1--E85. E14, E53 retired and E66/E69--E73/E75/E78/E79
unused (absent from the text); E28--E29 are described in the Experimental Setup but
their results section was cut.}\label{tab:experiment-index}\\
\toprule
\textbf{E\#} & \textbf{Description} & \textbf{Where} \\
\midrule
\endfirsthead
\toprule
\textbf{E\#} & \textbf{Description} & \textbf{Where} \\
\midrule
\endhead
\midrule
\multicolumn{3}{r}{\footnotesize\itshape continued on next page}\\
\endfoot
\bottomrule
\endlastfoot
E1 & Head-to-head symbolic vs.\ learned dynamics on the sprint world & Tab.~\ref{tab:main} \\
E2 & Synthesis reliability: five compile attempts per world & Tab.~\ref{tab:synthesis-app} \\
E4 & Head-to-head engine comparison, primary sprint result & Tab.~\ref{tab:main} \\
E9 & Configuration-search strategies on triage auto-tuning & Tab.~\ref{tab:tuning-app} \\
E10 & Head-to-head primary comparison, throughput and accuracy & Tab.~\ref{tab:main} \\
E11 & Multi-world fidelity replication, eight scripts per world & Tab.~\ref{tab:fidelity} \\
E12 & Trained MLP/1-NN baselines vs.\ the zero-transition program & Fig.~\ref{fig:learned} \\
E14 & \emph{Retired (absorbed); not in text} & --- \\
E16 & Cross-family synthesis replication (Llama, Gemma) & Tab.~\ref{tab:synthesis-app} \\
E18 & Filter-vs-repair decomposition of the verification gate & Tab.~\ref{tab:repair-app} \\
E19 & Scale-ladder vs.\ a stronger learned baseline, OOD collapse & Tab.~\ref{tab:ladder-app} \\
E20 & The complexity cliff: synthesis degrades past eight rules & Fig.~\ref{fig:complexity} \\
E21 & Stochastic dynamics via seed-threaded verified synthesis & \S\ref{sec:boundaries} \\
E22 & Planning: model-predictive lookahead through the program & Tab.~\ref{tab:planning} \\
E23 & Oracle-free error detection by ensemble disagreement & \S\ref{sec:boundaries} \\
E28 & \emph{Cut (setup only)}: atomic SWE-style repair suite & \S~Setup \\
E29 & \emph{Cut (setup only)}: staged latent-defect repair suite & \S~Setup \\
E37 & Equal-information induction: code beats learning OOD & Tab.~\ref{tab:induction} \\
E38 & The induction ceiling is identifiability, not capability & Fig.~\ref{fig:induction-scale} \\
E43 & Active induction: acting closes the identifiability gap & Fig.~\ref{fig:active-induction} \\
E53 & \emph{Retired (sheaf section cut); not in text} & --- \\
E61 & Verified code vs a trained world model on control (absorbed into E63) & Tab.~\ref{tab:bakeoff} \\
E63 & World-model bake-off: 6 runnable models $\times$ 2 domains & Tab.~\ref{tab:bakeoff} \\
E65 & Cross-species on MiniGrid: verified code vs.\ DreamerV3 / V-JEPA-2 & \S\ref{sec:minigrid} \\
E67 & Continuous control: cartpole-swingup vs.\ learned/perceptual models & \S\ref{sec:cartpole} \\
E74 & World-time compute: held-out lift vs.\ model size (diagnosis family) & \S\ref{sec:world-time} \\
E76 & World-count scaling of the world-time-compute lever & \S\ref{sec:world-time-scale} \\
E77 & Coding-world transfer of world-time compute (pass@5) & \S\ref{sec:world-time-scale} \\
E78b & Label-exactness ablation isolates the causal driver & \S\ref{sec:world-time-scale} \\
E80 & Real-data world-time compute on ARC-AGI (per-world TTT + ladder) & \S\ref{sec:world-time-arc} \\
E80 & Replication across List Functions, CLRS, Bongard-RWR (vision boundary) & \S\ref{sec:world-time-domains} \\
E80 & A descriptive regularity: LODO fit, tabular out-of-sample, learner-invariance & \S\ref{sec:world-time-law} \\
E81 & Frame-invariance across language frames (clean negative) & \S\ref{sec:frame-invariance} \\
E82 & The hybrid loop: verified world as generator + inference backstop & \S\ref{sec:hybrid-loop} \\
E83 & Real-data cross-world transfer on CLRS-Text (weak null) & \S\ref{sec:world-time-law} \\
E84 & Cross-world transfer on a real domain (List Functions); generalize to held-out worlds & \S\ref{sec:crossworld} \\
E85 & Trajectory vs.\ answer training unit on iterated worlds & \S\ref{sec:discussion} \\
\end{longtable}
\endgroup

\section{Detailed result tables}\label{app:tables}

\begin{table}[htb]\centering\small
\resizebox{\linewidth}{!}{%
\begin{tabular}{lcccc}
\toprule
Generator (family) & Runs & Verified acceptance (95\% CI) & Probe acc.\ of accepted & Mean synthesis time \\
\midrule
qwen2.5:7b (Qwen) & 15 & 100\% [0.80, 1.00] & 0.98 & --- \\
qwen2.5:3b (Qwen) & 15 & 100\% [0.80, 1.00] & 0.87 & --- \\
llama3.1:8b (Llama) & 9 & 100\% [0.70, 1.00] & 0.85 & 9s \\
gemma2:9b (Gemma) & 9 & 100\% [0.70, 1.00] & 1.00 & 4s \\
\bottomrule
\end{tabular}
}

\caption{E2 + E16: synthesis reliability by generator scale and family.
Qwen rows: five compilation attempts per world; Llama/Gemma rows: three per
world. Maximum four verify--repair iterations each; wall-clock includes all
iterations.}
\label{tab:synthesis-app}
\end{table}

\begin{table}[htb]\centering\small
\begin{tabular}{lccc}
\toprule
Regime & Acceptance & Probe acc.\ (accepted) & Probe acc.\ (all runs) \\
\midrule
Filter only (max\_iters=1) & 62\% & 0.62 & 0.39 \\
Filter + repair loop (max\_iters=4) & 100\% & 0.65 & 0.65 \\
\bottomrule
\end{tabular}

\caption{E18: filter-vs-repair decomposition of the full verification gate
(3B generator, eight paired high-temperature attempts per regime).}
\label{tab:repair-app}
\end{table}

\begin{table}[htb]\centering\small
\begin{tabular}{lccc}
\toprule
Engine & 1$\times$ & 10$\times$ & 100$\times$ \\
\midrule
\textbf{Synthesized code (0 transitions)} & 10/10 & 10/10 & 10/10 \\
Delta-MLP, 128h (10k transitions) & 5/10 & 0/10 & 0/10 \\
MLP, 64h (10k transitions) & 5/10 & 0/10 & 0/10 \\
1-NN memorizer (10k transitions) & 3/10 & 0/10 & 0/10 \\
LLM next-state (7B, 0 transitions) & 4/10 & 4/10 & 5/10 \\
\bottomrule
\end{tabular}

\caption{E19: exact transition accuracy on the branch-covering probe suite
across a 1$\times$/10$\times$/100$\times$ scale ladder.}
\label{tab:ladder-app}
\end{table}

\begin{table}[htb]\centering\small
\begin{tabular}{lccc}
\toprule
Strategy & Seeds solving & Mean trials to first solve & Mean best score \\
\midrule
random & 10/10 & 4 & 17.60 \\
tpe & 10/10 & 5 & 17.60 \\
random+refine & 10/10 & 4 & 17.60 \\
\bottomrule
\end{tabular}

\caption{E9: tuning-strategy comparison on the triage auto-configuration
problem; ten seeds per strategy, \TuningBudget-trial budget.}
\label{tab:tuning-app}
\end{table}

\section{The synthesized dynamics artifact}\label{app:figs}
A representative accepted sprint-world transition, synthesized by
\texttt{qwen2.5:7b} and verified by the full gate---the artifact whose
exactness underwrites Table~\ref{tab:main}:

\begin{tcolorbox}[policycard, cardcol=cardteal,
  title=Accepted \texttt{transition()} (verbatim)]
\begin{lstlisting}[style=housespecyamlbare]
def transition(state, action):
    next_state = dict(state)
    name = action["name"]
    if name == "ship" and next_state["backlog"] > 0:
        next_state["backlog"] -= 1
        next_state["shipped"] += 1
        next_state["debt"] += 1
        next_state["bugs"] += next_state["debt"] // 4
    elif name == "fix":
        next_state["bugs"] = max(0, next_state["bugs"] - 2)
    elif name == "refactor":
        next_state["debt"] = max(0, next_state["debt"] - 2)
    return next_state
\end{lstlisting}
\end{tcolorbox}

\section{Example tasks and failure cases}\label{app:examples}
The coding-world tasks pair a buggy function with hidden tests; e.g.\
\texttt{dedupe} returns \texttt{list(set(xs))} (order-destroying) and must
preserve first-seen order, and \texttt{balanced} counts parenthesis depth but
misses early-imbalance strings like \texttt{")("}. Failure modes observed in
the campaign: the 1.5B baseline resubmitted near-identical patches until the
attempt budget expired on tasks where its first reading of the spec was
wrong---precisely the local minimum that high-temperature candidate sampling
plus judge selection escapes; conversely, when several candidates passed, the
judge's pick was strongly order-sensitive (E13: raw choices matched under
order reversal on only \OrderConsistency{} of rounds) while its accuracy was
not---evidence that judge criteria and presentation order need auditing even
when outcomes look fine. In E7/E15, the rubric judge compressed mid-range
episodes toward similar scores (rank correlation \JudgeSpearman{} rather than
1.0, dropping to \RubricParaRho{} under paraphrase), consistent with known
LLM-judge coarseness~\cite{llmjudge2023}.

\section{Reproducibility}
One pipeline rebuilds every artifact: experiment scripts in
\texttt{experiments/} write \texttt{results/*.json} (seeds fixed in-source);
\texttt{python3 scripts/make\_paper\_assets.py} regenerates every figure,
table, and the \texttt{numbers.tex} macros from those JSON files; and
\texttt{make} in \texttt{paper/} builds this PDF. A continuous-integration
workflow runs the offline test suite, re-runs the asset pipeline, and fails if
\texttt{numbers.tex} would change, so the paper's numbers cannot silently drift
from the committed results. \texttt{experiments/MANIFEST.md} classifies every
experiment as deterministic-offline or Ollama-dependent and records its model and
provenance; \texttt{experiments/requirements.lock} pins the analysis stack
(Python, numpy, matplotlib). The \NumModels{} Ollama model snapshots that appear
across the runs are \ModelList{} (quantization \texttt{Q4\_K\_M}); platform,
versions, and timestamp are recorded in each result file. Deterministic-offline
experiments reproduce bit-for-bit; the Ollama-dependent runs reproduce
in distribution (Metal is not bit-deterministic even at fixed seed), with
dataset/model-pinned frozen artifacts where exact numbers are cited.

\clearpage

\bibliographystyle{plain}
\bibliography{refs}
\end{document}